%% file: main.tex
\documentclass{article} %
\PassOptionsToPackage{table,xcdraw}{xcolor} %
\usepackage{iclr2027_conference,times}

\usepackage[utf8]{inputenc} %
\usepackage[T1]{fontenc}    %
\usepackage{microtype}      %

\input{math_commands.tex}

\usepackage{hyperref}
\usepackage{url}
\usepackage[capitalize]{cleveref}

\usepackage{xspace}
\usepackage{multirow}
\usepackage{booktabs}
\usepackage[]{todonotes}

\usepackage{algorithm}
\usepackage{algpseudocode}
\usepackage{enumitem}
\usepackage{float}
\usepackage{fvextra} 

\usepackage[labelfont=bf, justification=justified,font=small]{caption}
\usepackage{xcolor} %

\input{macro}

\usepackage{amsthm}
\newtheorem{proposition}{Proposition}

\title{\methodname: Improving Self-Play for Long-Horizon Tasks with Decomposable Verifiable Rewards}

\author{Tianyu Liu\symbolthanks{2}{Work completed during an internship at Qualcomm AI Research}\, \rlap{$^1$}, \; 
Davide Belli\rlap{$^2$}, \; 
Bence Major\rlap{$^2$}, \; 
Arash Behboodi\rlap{$^2$}, \; 
Tribhuvanesh Orekondy\rlap{$^2$} \\[0.5em]
{$^1$} ETH Zurich \quad {$^2$} 
Qualcomm AI Research\symbolthanks{3}{Qualcomm AI Research is an initiative of Qualcomm Technologies, Inc.}
}

\newcommand{\methodname}{{Explorer-Zero}\xspace}

\begin{document}
\iclrfinalcopy

\maketitle

 \begin{abstract}
  Training LLM agents on long-horizon interactive tasks with reinforcement learning typically relies on human-annotated tasks and reliable verification, both expensive to construct for stateful environments.
  Self-play offers an alternative, but must generate achievable, compositional tasks and provide feedback beyond binary outcomes.
  We propose \textbf{\methodname}, a self-play framework that \emph{jointly generates tasks and fine-grained, verifiable rewards through environment interaction}.
  A learned proposer explores the environment, constructs tasks, and records intermediate and final states as verification targets.
  These checkpoints reward partial completion, while alternating proposer--solver training adapts the generated curriculum.
  We show theoretically that subtask rewards reduce collapse among unsuccessful rollouts when they complete different numbers of subtasks.
  Without human-annotated training tasks, \methodname outperforms the evaluated self-play baselines across ALFWorld, ScienceWorld, and AppWorld; on AppWorld, our 30B agent reaches $59.67\%$ task success, approaching $62.65\%$ for RL on human-annotated tasks.
  Further analyses support task-relevant composition and benefits on complex tasks, while self-play initialization improves subsequent human-data RL across all three benchmarks.
  \end{abstract}

\section{Introduction}
LLM agents are increasingly applied to open-ended, long-horizon tasks, from coding \citep{jimenez2024swebench} and tool use \citep{yao2024taubench} to embodied environments \citep{shridhar2021alfworld} and application control \citep{trivedi2024appworld}.
The dominant training recipe is reinforcement learning with verifiable rewards (RLVR), but it depends on tasks that are achievable, diverse, and paired with reliable success checks \citep{lambert2024tulu3,guo2025deepseek}.
Creating such supervision for stateful environments is expensive: task designers must understand what the current state permits, specify a meaningful change to it, and define how to verify that change.
Human task construction therefore limits how broadly agentic RL can scale.

Self-play can remove this bottleneck by training a proposer to generate tasks for a solver.
It has been particularly effective when tasks and success checks can be produced from static artifacts: game rules determine the winner \citep{silver2017mastering,sukhbaatar2018intrinsic}, execution or formal checking validates code \citep{zhao2025absolute,wilf2026propose}, retrieved evidence supplies answers to search questions \citep{yue2026drzero,lu2026searchselfplay}, and structured specifications define correct tool calls \citep{acikgoz2026toolr0}.
Long-horizon interaction in a stateful environment is different: the proposer often cannot determine from the initial observation which goals are reachable, and a final binary success check discards the progress made toward an unsuccessful task.

Our approach is motivated by two observations.
First, long-horizon agentic tasks are often \emph{compositional}: their completion requires a sequence of meaningful state changes.
For instance, in embodied scenarios, an agent must locate an item before it can pick it up and move it to a destination.
Generating each task in one shot overlooks this structure; a proposer that acts in the environment can instead discover feasible components and compose them into a grounded task.
Second, the delayed outcomes of long-horizon tasks make sparse rewards particularly limiting, while fine-grained feedback can provide a substantially more useful learning signal \citep{ng1999policy,arjonamedina2019rudder,lightman2024verify}.
In group-normalized RL, this can cause \emph{reward collapse}: a binary reward assigns the same value to sampled trajectories, so their advantages vanish despite meaningful differences in progress \citep{razin2025what}.

We introduce \methodname, a self-play framework in which a learned proposer explores the environment and generates a natural-language task grounded in its trajectory.
As it acts, the proposer declares meaningful subtasks and saves the corresponding environment states; its final state provides the task-level success check.
This design has two benefits.
First, exploration lets the proposer build compositional tasks from demonstrated state transitions, producing complex training problems that better reflect real long-horizon use than isolated goals.
Second, the saved intermediate states become verifiable subgoal checkpoints: failed solver trajectories receive fine-grained rewards according to the subtasks they complete, reducing the frequency with which binary rewards collapse and discard partial progress.
Because the proposer itself is trained alongside the solver, the generated curriculum can evolve as the solver improves without relying on human-written tasks or a prompted frontier model.\looseness=-1

Our main contributions are:
  \begin{enumerate}[leftmargin=*,nosep]
      \item \textbf{Self-play that jointly generates compositional tasks and supervision.}
      We introduce a learned proposer that explores stateful environments, constructs compositional tasks, and records verifiable subtask checkpoints.
      These checkpoints provide fine-grained solver rewards, enabling self-play without human-authored training tasks or learned judges.

      \item \textbf{Theoretical and empirical analysis of subtask supervision.}
      We derive conditions under which subtask rewards preserve a learning signal even when all sampled trajectories fail to complete the task.
      Our empirical analyses show that the subtask incentive encourages task-relevant compositions, with downstream gains concentrated on tasks requiring multiple dependent
      operations.\looseness=-1

      \item \textbf{Effective agent training with and without human-annotated tasks.}
      Across ALFWorld, ScienceWorld, and AppWorld, \methodname outperforms the evaluated self-play baselines without human-annotated training tasks and improves final performance
      when used to initialize RL on human-annotated tasks.
      Controlled ablations quantify the benefits of interactive proposal and subtask supervision. \looseness=-1
  \end{enumerate}

Without human-annotated training tasks, \methodname outperforms Self-Challenging, the strongest evaluated self-play baseline, by $18.67$ points on ALFWorld and $7.23$ on ScienceWorld with 4B agents. On AppWorld, the 30B agent reaches $59.67$ TGC, exceeding this baseline by $7.74$ points and approaching human-data RL ($62.65$). Self-play initialization also improves subsequent human-data RL across all three environments.

\section{Related Work}
\label{sec:related}

\myparagraph{Self-play with verifiable tasks.}
Self-play has produced superhuman game-playing policies and automatic curricula in embodied control \citep{silver2017mastering,sukhbaatar2018intrinsic,openai2021asymmetricselfplayautomaticgoal}.
Recent proposer--solver methods generate code problems checked by execution or formal verification \citep{zhao2025absolute,wilf2026propose}, while R-Zero uses a challenger--solver loop to generate verifiable reasoning problems \citep{huang2026rzero}.
Related approaches generate questions grounded in documents or retrieved evidence \citep{yang2026spell,yue2026drzero,lu2026searchselfplay} and structured tool-call problems \citep{acikgoz2026toolr0}; J-Zero instead uses a learned judge for open-ended tasks \citep{chu2026jzero}.
GenEnv and SPADE extend self-play by generating executable environments and adapting their difficulty to the solver \citep{guo2025genenv,liu2026spade}.
These approaches rely on known answers, automatic checks, or model-generated environments, and therefore do not address the central challenge in a stateful target environment: discovering reachable, compositional goals through interaction and retaining a learning signal when the solver cannot yet complete them.%

\myparagraph{Interaction-grounded task generation.}
Several methods generate training tasks from environment interaction.
Learn-by-Interact converts collected trajectories into instructions \citep{su2025learnbyinteract}, while Self-Challenging Language Model Agents \citep{zhou2025selfchallenging} has a tool-using challenger produce an instruction, executable verifier, example solution, and failure cases.
AutoPlay \citep{ramrakhya2025autoplay} verifies explored UI tasks with a multimodal judge and filters them by executor success.
AgentSynth \citep{xie2025agentsynth}  composes executor-performed subtasks into longer computer-use tasks, but does not reuse them as reward checkpoints.
In these methods, task generation is performed by fixed, prompted models rather than learned jointly with the solver.
Moreover, their interaction trajectories are not reused as intermediate reward checkpoints, so partial progress is not preserved as supervision.
Relatedly, hindsight experience replay relabels trajectories with reached goals \citep{andrychowicz2017hindsight}, and concurrent State2State \citep{lei2026state2state} turns states collected largely through random exploration into state-matched goals for mid-training.
Unlike methods centered on individual reached states, \methodname trains the proposer to compose a natural-language task from multiple demonstrated transitions and uses the same states to verify intermediate progress.%

\myparagraph{Fine-grained supervision for long-horizon RL.}
Fine-grained feedback is well established in RL: reward shaping augments sparse task rewards with intermediate signals \citep{ng1999policy}, RUDDER \citep{arjonamedina2019rudder} attributes delayed rewards to consequential earlier decisions, and process supervision applies intermediate feedback to language-model reasoning \citep{lightman2024verify}.
Theory further shows that even accurate rewards provide little learning signal when they fail to distinguish outputs from the current policy \citep{razin2025what}.
To address this problem, self-play methods typically steer proposers toward tasks with intermediate solve rates \citep{zhao2025absolute,huang2026rzero,yue2026drzero,guo2025genenv}.
SPADE \citep{liu2026spade} instead uses the performance gap with and without a privileged hint to guide environment generation.
These approaches improve task selection rather than derive fine-grained solver rewards from task generation.
Outside self-play, GiGPO and milestone-guided shaping provide finer feedback on predefined agent tasks through matched states or hand-designed progress rules \citep{feng2025gigpo,wang2026milestone}.
\methodname brings fine-grained supervision into self-play: the same proposer trajectory generates the task and the verifiable checkpoints used to reward partial completion.

\section{Methodology}
\label{sec:method}

We introduce \methodname, a proposer--solver self-play framework for long-horizon stateful environments. \looseness=-1

\myparagraph{Problem setup and notation.}
An agent with \defn{policy} $\policy_{\param}$ interacts with a stateful \defn{environment} $\env$: given a \defn{task} $\task$, it produces a \defn{trajectory} $\traj = (\actionsym_1, \obs_1, \dots, \actionsym_T, \obs_T)$ of \defn{actions} $\actionsym_t$ and \defn{observations} $\obs_t$, traversing \defn{environment states} $\state_0, \dots, \state_T$.

\myparagraph{Standard RLVR and its data-curation bottleneck.} 
Standard RLVR trains an agent with policy-gradient methods to maximize expected reward over a fixed task distribution: 
$\max_{\param}\; \E_{\task \sim \taskdist,\, \traj \sim \policy_{\param}(\cdot \mid \task)} \left[\reward(\task,\traj)\right]$.
A verifier assigns a typically binary reward $\reward(\task,\traj)\in\{0,1\}$ using an executable test, answer check, or state-based predicate. 
Building this training distribution requires human-authored tasks, contexts, and reliable success checks. 
Examples include initial world states and rule-based checks in ALFWorld and ScienceWorld \citep{shridhar2021alfworld,wang2022scienceworld}, and initial database states and unit tests in AppWorld \citep{trivedi2024appworld}. 
This curation is expensive because tasks must be both achievable and challenging, while verification must remain robust to reward hacking. 
For example, \citet{trivedi2024appworld} report five months of human effort for data generation.\looseness=-1

\begin{figure}[t]
  \centering
  \includegraphics[width=0.90\linewidth]{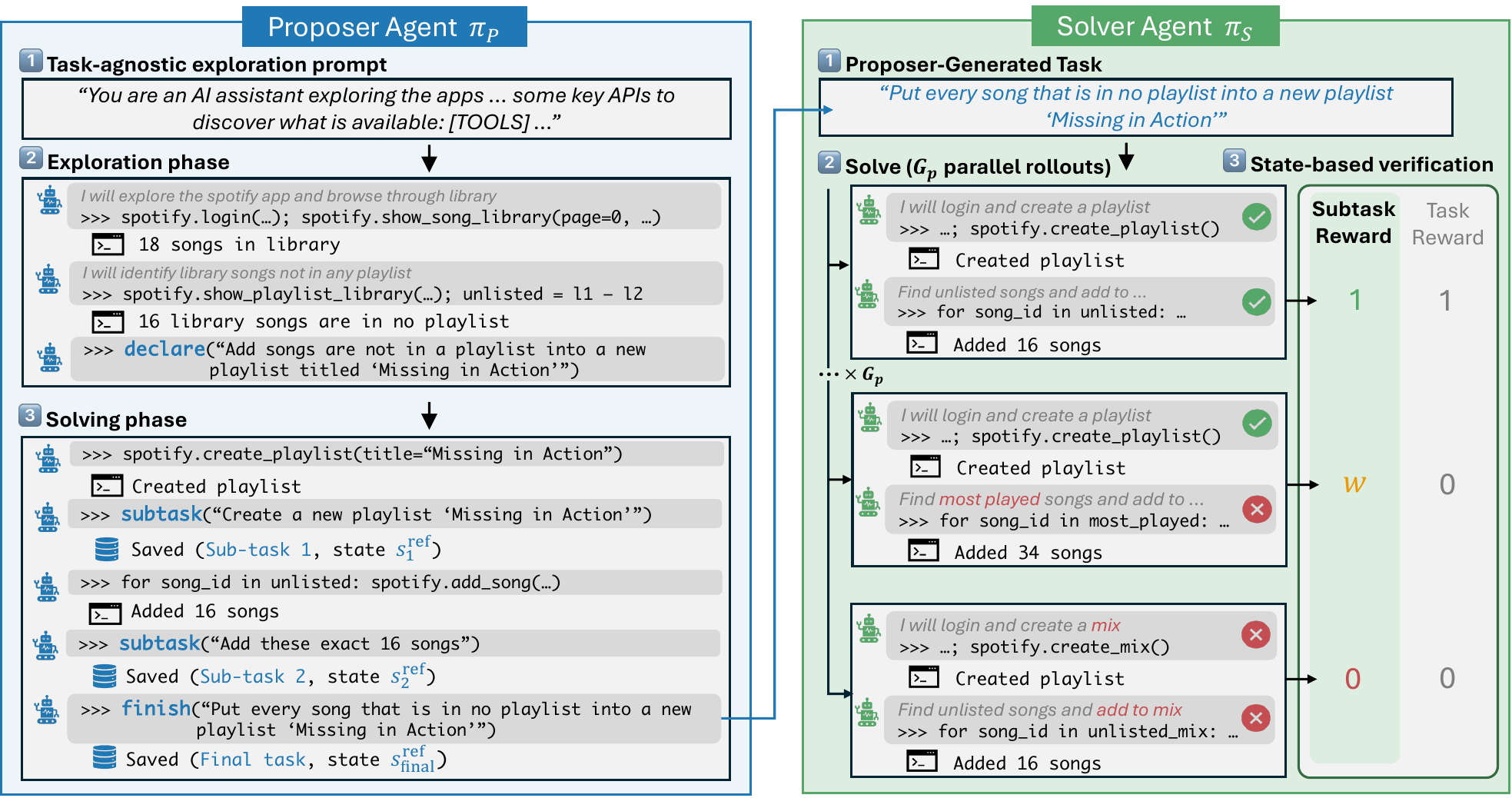}
  \caption{
    Overview of \methodname.
    \textbf{Left:} the proposer $\proposerpolicy$ explores the environment, declares an initial task intent ($\declaretasktool$), completes subtasks ($\subfinishtool$) and the final task $\task$ ($\finishtool$). Each $\subfinishtool$ and $\finishtool$ call snapshots the environment as a reference state, yielding $\stateref_1, \dots, \stateref_{\numsubtasks}$ and $\finalstateref$.
    \textbf{Right:} the solver $\solverpolicy$'s traversed states are matched against the reference states to compute $\solverreward$, and the resulting solve rate determines $\proposerreward$.\looseness=-1
  }
  \label{fig:illustration}
\vspace{-3em}
\end{figure}

\myparagraph{Our approach: High-level idea.}
\methodname replaces the fixed task distribution with tasks generated by a
\defn{proposer} $\proposerpolicy$.
The proposer interacts with the environment to generate a task and state-based verification artifacts.
A \defn{solver} $\solverpolicy$ is trained on the accepted tasks.
The two policies are optimized in alternating phases, allowing the task distribution to evolve with the solver. Our method is illustrated in \cref{fig:illustration}, and the full procedure is shown in \cref{alg:selfplay}.

\subsection{Proposer: Interactive Generation of Compositional Tasks}
\label{sec:proposer}

During proposer training, the solver is frozen.
We describe how the proposer generates tasks, verification artifacts, and its training reward.

\myparagraph{Proposer: Main idea.}
Starting from $\state_0$, the proposer at the end of a successful rollout generates
$\left(
    \task,\finalstateref,
    \{(\subtask_k,\stateref_k)\}_{k=1}^{\numsubtasks}
\right)$,
where $\task$ is the final task, $\finalstateref$ is its reference outcome, and
$(\subtask_k,\stateref_k)$ are optional intermediate checkpoints.
All reference states are reached during the proposer trajectory, grounding the task in a demonstrated outcome.
Here, the final task $\task$ and subtask descriptions $\subtask_k$ are in natural language (tokens generated by the LLM policy) and the states $\stateref_k$ correspond to snapshots of the environmental state.

\myparagraph{Phases: Exploration to task generation.}
The proposer interacts with the environment for at most $T_{\mathrm{P}}$ steps.
Proposal proceeds in two phases. 
During the \defn{exploration phase}, the proposer examines the environment over multiple turns and identifies a candidate task. 
During the \defn{solving phase}, the proposer attempts to complete its own task. 
It may label meaningful subtasks as they are completed and finally submits the final task description. 
We denote the complete interaction by the \defn{proposer trajectory} $\proptraj$.
This separation serves two purposes. 
Exploration allows the proposer to discover what the current environment state permits, while solving demonstrates that the proposed outcome is reachable. 
It also allows the final task description to reflect information obtained during interaction rather than only the initial observation.

\myparagraph{Verification artifacts. }
Generating a task description alone is insufficient for RLVR because the solver also requires a reliable task-specific verification signal. 
To enable this, we use environment states reached by the proposer as verification artifacts. 
When the proposer completes its task, we snapshot the resulting environment state as the final reference state $\finalstateref$. 
Given a solver trajectory with terminal state $\state_T$, task completion can then be checked by comparing $\state_T$ with $\finalstateref$. 
The proposer trajectory therefore provides both the task and an executable state-based verification target without requiring a task-specific predicate, model-based judge, or model-written test. 
We evaluate state equality through changes relative to the common initial state $\state_0$. 
A solver state $\state$ matches a reference state $\stateref$, written $\state \simeq \stateref$, if $\state$ reproduces the task-relevant changes exhibited by $\stateref$ relative to $\state_0$. 
The concrete matching rules for each environment are described in \cref{sec:experiments}.

\myparagraph{Compositionality.}
The proposer trajectory provides more than a final verification target. 
As the proposer solves its task, it can identify meaningful intermediate outcomes and snapshot their corresponding states. 
Each declared subtask $\subtask_k$ is therefore paired with a reference state $\stateref_k$ reached after completing that component. These intermediate states expose the compositional structure of the generated task. 
Instead of predicting a distant goal in one step, the proposer can establish one outcome, continue interacting from the resulting state, and construct later components that depend on earlier ones. 
The final task describes the cumulative outcome of these demonstrated transitions. 
For simplicity, we use the declared subtasks as an unordered checklist of verification conditions, without explicitly modeling dependencies between them.
A solver may satisfy them in any order, provided that their state conditions are reached. 
The proposer chooses the number of subtasks $\numsubtasks$ and when $\numsubtasks=0$, it generates only the final task and its reference state.

\myparagraph{Proposer's reward.}
For each proposal, we sample $\numscore$ trajectories from the frozen solver and estimate
$\solverate(\task)
=
\frac{1}{\numscore}
\sum_{i=1}^{\numscore}
\reward(\task,\traj_i)$,
where $\reward(\task,\traj_i)\in\{0,1\}$ indicates final-task completion using state-based verification.
The proposer reward is
\begin{equation}
\label{eq:proposer-reward}
    \proposerreward(\task,\proptraj)
    =
    \begin{cases}
        \difficultyreward(\task)
        + \subtaskreward(\task,\proptraj)
        + \diversityreward(\proptraj),
        & \textnormal{if the proposal is valid},\\
        0, & \textnormal{otherwise}.
    \end{cases}
\end{equation}
The difficulty reward is:
\begin{equation}
    \difficultyreward(\task)
    =
    1-\left|2\solverate(\task)-1\right|
\end{equation}
which is maximized at a solve rate of $0.5$ and penalizes tasks that are consistently solved or consistently failed.
The remaining terms encourage intermediate structure and diverse interaction:
\begin{align}
    \label{eq:subtaskreward}
    \subtaskreward(\task,\proptraj)
    &= \subtaskweight\numsubtasks,\\
    \diversityreward(\proptraj)
    &= \diversityweight
       \left|\actiontypes(\proptraj)\right|,
\end{align}
where $\numsubtasks$ is the number of declared subtasks, $\actiontypes(\proptraj)$ is the set of distinct proposer action types, and $\subtaskweight,\diversityweight>0$ are weighting coefficients.

\myparagraph{Implementation details.}
The proposer starts from the same state $\state_0$ later used for solver rollouts.
Each proposer GRPO group shares one initial state (an ALFWorld or ScienceWorld environment instance, or an AppWorld persona).
Its prompt contains the environment interface and encourages broad exploration.
In addition to environment actions, it has three control tools:
(i) $\declaretasktool(\texttt{task\_description})$ ends exploration and declares a candidate task;
(ii) $\subfinishtool(\texttt{subtask\_description})$ records a completed subtask and its reference artifact;
(iii) $\finishtool(\texttt{task\_description})$ submits the final task and records $\finalstateref$.
For state-changing outcomes, the reference artifact is the state delta relative to $\state_0$.
For observational subtasks without a persistent state change, it also records the action or API call that produced the relevant observation and its returned value, which the solver must reproduce.
We define action types as command verbs in ALFWorld and ScienceWorld, and as applications in AppWorld.
The coefficients $\subtaskweight,\diversityweight>0$ control the two incentives.
Two subtasks are redundant if their state deltas are identical and their descriptions have word-level Jaccard similarity above $0.6$. The subtask term in \cref{eq:subtaskreward} is applied only to non-redundant checkpoints.

\subsection{Solver: Training a Policy to Solve}
\label{sec:solver}

During solver training, the proposer is frozen.
The solver is trained on accepted proposer-generated tasks using final-task and subtask verification.

\myparagraph{Task generation and filtering.}
At each solver-training iteration, the frozen proposer samples tasks and their associated verification artifacts from a set of initial world states. 
A generated example consists of an initial state $\state_0$, a task description $\task$, a final reference state $\finalstateref$, and zero or more subtask reference states $\stateref_1,\dots,\stateref_{\numsubtasks}$. 
Before using an example for solver training, we apply a rule-based validity filter. 
We retain a task only if: 
(i) the proposer submits a natural-language task through $\finishtool$; 
(ii) the task contains no redundant subtasks; and 
(iii) every application or object named in the task description appears in the proposer trajectory. 
These checks remove malformed proposals, repeated checkpoints, and instructions that are unsupported by the demonstrated interaction. 
Tasks rejected by the filter are not included in the solver training distribution.

\myparagraph{Task and rollout.}
For each accepted task, the environment is reset to the proposer's initial state $\state_0$.
The solver observes only the task description and environment interface, not the proposer trajectory nor verification artifacts.
The solver generates
$\solvtraj=(\actionsym_1,\obs_1,\dots,\actionsym_T,\obs_T)$.
We retain its visited states $\gS(\solvtraj)=\{\state_0,\dots,\state_T\}$,
so that a subtask remains verifiable even if its state is later modified.

\myparagraph{State-based verification and rewards.}
The final-task reward compares the solver's terminal state with the final reference state:
\begin{equation}
\label{eq:task-reward}
    \reward(\task,\solvtraj)
    =
    \indicator\left\{
        \state_T \simeq \finalstateref
    \right\}.
\end{equation}
For $\numsubtasks>0$, the subtask reward is
\begin{equation}
\label{eq:subgoal-reward}
    \subgoalreward(\task,\solvtraj)
    =
    \frac{1}{\numsubtasks}
    \sum_{k=1}^{\numsubtasks}
    \indicator\left\{
        \exists\,\state\in\gS(\solvtraj):
        \state\simeq\stateref_k
    \right\}.
\end{equation}
We set $\subgoalreward(\task,\solvtraj)=0$ when $\numsubtasks=0$. The solver reward is
\begin{equation}
\label{eq:solver-reward}
    \solverreward(\task,\solvtraj)
    =
    \begin{cases}
        1,
        & \textnormal{if }\reward(\task,\solvtraj)=1,\\
        \subgoalreward(\task,\solvtraj),
        & \textnormal{otherwise}.
    \end{cases}
\end{equation}
Thus, successful trajectories receive the maximum reward, while unsuccessful trajectories are distinguished by verified subtask completion.
The solver is updated with a group-based policy-gradient objective over multiple rollouts per task.

\subsection{Subtask Rewards Reduce Reward Collapse}
  \label{sec:reward-collapse}

We now examine how the reward in \cref{eq:solver-reward} can preserve a learning signal on tasks that the solver rarely completes. With binary outcome rewards, unsuccessful trajectories all receive zero reward, even when they make different amounts of progress. GRPO computes advantages by normalizing rewards within a group of $\groupsize \ge 2$ i.i.d.\ rollouts (\cref{app:prelim}). When all rollouts receive the same reward, every reward-normalized advantage is zero\footnote{We adopt the standard convention that the advantage is set to zero when the group std vanishes.}, and the group provides no reward-driven policy-gradient signal. We refer to this event as \defn{reward collapse}.

To quantify how subtask rewards affect collapse, let $p(\task) = \Pr_{\solvtraj \sim \solverpolicy(\cdot \mid \task)} [\reward(\task, \solvtraj) = 1]$ denote the current solver's success probability, estimated empirically by $\solverate(\task)$ given a task $\task$.

  \begin{proposition}[Subtask rewards reduce reward collapse]
  \label{prop:zerograd}
  Under the binary task reward, a group collapses with probability
  $p^{\groupsize} + (1-p)^{\groupsize}$.
  Conditional on all sampled rollouts failing, binary rewards collapse with probability one.
  Under \cref{eq:solver-reward}, conditional on all rollouts failing, the collapse probability is
  $\sum_{j=0}^{\numsubtasks} \tilde{q}_j^{\,\groupsize}$,
  where $\tilde{q}_j$ is the probability that a failed rollout completes exactly $j$ subtasks.
  This probability is strictly below one whenever at least two subtask-completion counts have positive probability among failed rollouts.
  \end{proposition}

  The proof and an analysis of the effect of group size are given in \cref{app:proof}.
  Larger groups can reduce binary-reward collapse by increasing the chance of sampling a successful trajectory, but become costly on difficult tasks: as $p \to 0$, keeping the collapse probability below a fixed threshold requires $\groupsize$ to grow on the order of $1/p$.
  Higher sampling temperatures may likewise improve the chance of success through exploration, but neither sampling choice changes the fact that binary rewards assign identical values to all failed trajectories.
  Subtask rewards complement these choices by making differences in partial completion available for learning: even when every rollout fails, completing different numbers of subtasks yields nonzero normalized advantages.

\section{Experiments}
\label{sec:experiments}

\subsection{Experimental Setup}
\label{sec:experimental-setup}

\myparagraph{Benchmarks and evaluation.}
  We evaluate on ALFWorld \citep{shridhar2021alfworld}, a text-based household environment; ScienceWorld \citep{wang2022scienceworld}, a text-based science environment; and AppWorld \citep{trivedi2024appworld}, an ecosystem of API-controlled applications.
  Evaluation follows benchmark success criteria, not proposer-generated training checkpoints.
  Unless stated otherwise, we report task success rates on ALFWorld's unseen validation split and ScienceWorld's test split.
  On AppWorld's \texttt{test\_normal} split, we report \defn{task goal completion} (TGC), the percentage of tasks passing all benchmark tests, and \defn{scenario goal completion} (SGC), the percentage of scenarios with all task variants completed.
  Tables report means and standard deviations over four sampled evaluations per task.\looseness=-1

\myparagraph{Training-time verification.}
To compute rewards on self-generated training tasks, we instantiate the reference-state verification rule of \cref{sec:solver} for each environment.
In AppWorld, we compare changes in applications that changed in the reference trajectory and are mentioned in the task. %
In ALFWorld and ScienceWorld, we compare object types while ignoring instance indices, so manipulating interchangeable instances such as \texttt{apple 1} and \texttt{apple 2} is equally valid.

\myparagraph{Models and training.}
Self-play uses environment resets without human-authored training tasks, demonstrations, or reward annotations.
The proposer and solver are initialized from Qwen3-4B on ALFWorld and ScienceWorld and from Qwen3-4B-Instruct-2507 on AppWorld, where we additionally evaluate Qwen3-30B-A3B-Instruct-2507 \citep{yang2025qwen3}. We focus the larger-model evaluation on AppWorld, whose multi-application workflows allow us to test whether the benefits of compositional self-play persist with a stronger base agent.
We train with GRPO; hyperparameters and interaction budgets are given in \cref{app:details}.

\myparagraph{Baselines.}
  We compare two self-play approaches: Self-Challenging Agents \citep{zhou2025selfchallenging}, whose frozen proposer generates tasks and executable verifiers, and R-Zero \citep{huang2026rzero}, whose learned proposer generates tasks verified through solver consensus.
  Our R-Zero implementation uses the majority final state across solver rollouts as the pseudo-reference state.
  Additional comparisons include the zero-shot base model, in-context learning with $1$--$4$ human demonstrations, and RL on the full human-annotated training set as a reference for training with curated supervision.

\begin{table}[t]\centering\small
\caption{Experimental results on ALFWorld, ScienceWorld, and AppWorld. We report success rates on ALFWorld (unseen validation split) and ScienceWorld (test split), and TGC and SGC (task and scenario goal completion) on AppWorld (test normal split). We report means and standard deviations over four samples per task prompt.}
\label{tab:main}
\setlength{\tabcolsep}{3pt}%
\begin{tabular}{@{}lccccccc@{}}
\toprule
                             & {\#annotation}       & ALFWorld       & ScienceWorld & \multicolumn{4}{c}{AppWorld}     \\ \cmidrule(l){3-3}\cmidrule(l){4-4}\cmidrule(l){5-8}
                             &                               & \multicolumn{1}{c}{Qwen3-4B}  & \multicolumn{1}{c}{Qwen3-4B} & \multicolumn{2}{c}{Qwen3-4B-inst} & \multicolumn{2}{c}{Qwen3-30B-A3B-inst} \\ \cmidrule(lr){3-3}\cmidrule(lr){4-4}\cmidrule(lr){5-6}\cmidrule(l){7-8}
                             &                               &                &                & TGC          & SGC          & TGC          & SGC          \\ \midrule
\multicolumn{8}{c}{\cellcolor[HTML]{EFEFEF} \textit{Without} human-annotated training data}            \\ \midrule
    Baseline                     & n=0                           & \meanstd{17.14}{5.7}           & \meanstd{\phantom{0}7.03}{0.5}          & \meanstd{14.03}{5.1}         & \meanstd{\phantom{0}3.95}{3.2}         & \meanstd{25.15}{5.2}         & \meanstd{\phantom{0}8.04}{7.5}         \\ \midrule
In-Context          & n=1                           & \meanstd{19.28}{7.7}           & \meanstd{11.40}{0.4}          & \meanstd{12.28}{3.7}         & \meanstd{\phantom{0}6.58}{4.6}         & \meanstd{28.95}{5.0}         & \meanstd{15.79}{7.7}         \\
                             & n=2                           & \meanstd{18.57}{4.7}           & \meanstd{\phantom{0}9.50}{0.4}          & \meanstd{15.35}{4.0}         &  \meanstd{\phantom{0}1.32}{1.3}         & \meanstd{27.63}{4.8}         & \meanstd{11.84}{6.2}         \\
                             & n=3                           & \meanstd{33.57}{3.3}           & \meanstd{18.10}{0.7}          &\meanstd{14.91}{3.9}         & \meanstd{\phantom{0}3.95}{2.9}         & \meanstd{26.32}{5.3}         & \meanstd{14.47}{7.2}         \\
                             & n=4                           & \meanstd{30.71}{7.1}           & \meanstd{18.50}{0.3}          & \meanstd{13.16}{3.8}         & \meanstd{\phantom{0}2.63}{2.6}         & \meanstd{31.11}{0.2}         & \meanstd{17.11}{7.9}         \\ \midrule
R-Zero         & n=0                           & \meanstd{24.28}{1.5}           & \meanstd{24.76}{0.9}          & \meanstd{17.41}{5.8}         & \meanstd{\phantom{0}5.80}{1.6}         & \meanstd{47.32}{1.8}         & \meanstd{25.00}{2.7}         \\
Self-Challenging         & n=0                           & \meanstd{66.40}{3.5}           & \meanstd{35.69}{0.7}          & \meanstd{32.20}{3.2}         & \meanstd{16.25}{2.5}         & \meanstd{51.93}{1.3}         & \meanstd{32.25}{2.3}         \\
\textbf{Ours}         & n=0                           & \meanstd{\textbf{85.07}}{1.6}           & \meanstd{\textbf{42.92}}{0.7}          & \meanstd{\textbf{38.39}}{2.4}         & \meanstd{\textbf{16.96}}{1.9}         & \meanstd{\textbf{59.67}}{3.1}         & \meanstd{\textbf{39.29}}{3.2}         \\ \midrule
\multicolumn{8}{c}{\cellcolor[HTML]{EFEFEF} \textit{With} human-annotated training data} \\ \midrule
    RL    & full                          & \meanstd{95.71}{0.9}           & \meanstd{49.63}{0.4}          & \meanstd{46.38}{3.2}         & \meanstd{25.00}{2.4}          & \meanstd{62.65}{1.8}         & \meanstd{38.39}{3.2}         \\  
RL + \textbf{Ours}     & full                           & \meanstd{\textbf{98.57}}{0.9}           & \meanstd{\textbf{63.26}}{0.4}          & \meanstd{\textbf{50.00}}{2.0}         & \meanstd{\textbf{32.14}}{3.0}         & \meanstd{\textbf{69.20}}{4.6}         & \meanstd{\textbf{53.12}}{3.0}         \\  \bottomrule
\end{tabular}
\end{table}

\begin{table}[t]\centering\small
\caption{Ablation of \methodname components. Results on ALFWorld are on valid-seen, and results on AppWorld are with Qwen3-30B-A3B-Instruct-2507.}
\label{tab:ablation}
\begin{tabular}{@{}cccccc@{}}
\toprule
                           &                      & ALFWorld & ScienceWorld & \multicolumn{2}{c}{AppWorld } \\ \cmidrule(l){5-6}
                           &                      &          &          & TGC       & SGC       \\ \midrule
\multirow{4}{*}{Self-play} & one-shot proposer    & \meanstd{59.29}{9.4}     & \meanstd{\phantom{0}4.80}{2.0}     &   \meanstd{29.46}{4.7}      & \meanstd{16.96}{2.0}      \\
                           & interactive proposer  & \meanstd{65.71}{2.3}     & \meanstd{24.70}{1.2}     & \meanstd{45.83}{1.9}    & \meanstd{33.04}{2.7}      \\
                           & + subtask            & \meanstd{\textbf{84.29}}{1.5}     & \meanstd{\textbf{42.92}}{0.7}     &  \meanstd{\textbf{59.67}}{3.1}    & \meanstd{\textbf{39.29}}{3.2}      \\
                           \bottomrule
\end{tabular}
\vspace{-1.0em}
\end{table}

\subsection{Main Results and Ablations}
\label{sec:main-results}

\myparagraph{Comparison with self-play baselines.}
In \cref{tab:main}, \methodname outperforms both self-play baselines across all three environments without human-annotated training tasks.
Relative to Self-Challenging Agents (the stronger baseline) it improves ALFWorld performance by $18.67$ points and ScienceWorld performance by $7.23$ points.
On AppWorld, the TGC gains are $6.19$ points with the 4B model and $7.74$ with the 30B model, showing that the advantage persists at both model sizes.

\myparagraph{Comparison with human-data RL.}
RL on human-annotated tasks remains stronger on the primary task-level metrics, but the gap is particularly small for the 30B AppWorld agent: \methodname reaches $59.67$ TGC compared with $62.65$ for RL on the full training set.
This result shows that self-generated tasks and verification can support performance close to human-data RL in this setting, without using its task annotations.

\myparagraph{Combining \methodname and RL.}
We further evaluate self-play as a pretraining stage for RL on human-annotated tasks, denoted by RL + Ours in \cref{tab:main}.
We first train the solver with \methodname for $200$ steps, then continue training on human-annotated tasks for $150$ RL steps using the same hyperparameters as the RL-only baseline.
The combined protocol improves over RL alone across all three environments, including gains of $13.63$ points on ScienceWorld and $6.55$ TGC points on AppWorld with the 30B model.
These results support self-play as a useful initialization for agentic RL even when human-annotated tasks are available.

\myparagraph{\methodname ablations.}
\Cref{tab:ablation} compares three variants.
The \emph{one-shot proposer} generates tasks from the environment description and uses the most common final state across solver attempts as the reference outcome for verification.
The \emph{interactive proposer} grounds tasks and their reference outcomes in an executed trajectory, but rewards the solver only for final task completion.
The full method additionally declares subtasks and rewards intermediate progress.
Moving from one-shot to interactive proposal improves performance in all three environments, showing the value of grounding both task generation and verification in interaction.
Adding subtask supervision then yields a further $18.58$ points on ALFWorld, $18.22$ on ScienceWorld, and $13.84$ TGC points on AppWorld.
Subtask supervision thus adds substantial gains beyond interactive task generation.
This ablation measures the combined benefit of changing task construction and solver rewards; it does not isolate reward granularity alone. Training curves are included in \cref{app:curves}.

\section{Analysis}
\label{sec:empirical-analysis}

\begin{figure}[!t]
  \centering
  \includegraphics[width=\linewidth]{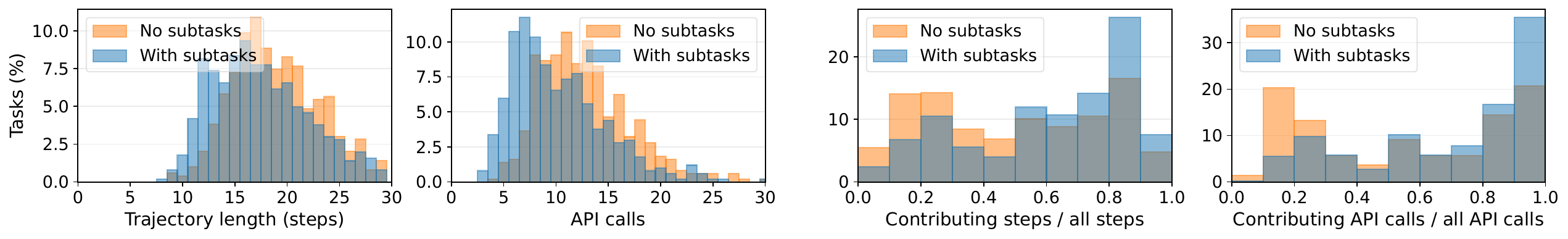}
  \caption{
    Task relevance of AppWorld proposer trajectories with and without subtask supervision.
    \textbf{Left two:} Trajectory lengths are similar across the two variants, while the proposer with subtasks makes fewer API calls on average.
    \textbf{Right two:} With subtasks, a larger fraction of steps and API calls contributes to the final task. We identify contributing interactions using backward program slicing \citep{weiser1984program}: they either produce target-state changes or support those changes through Python data flow.
    }
  \label{fig:analysis2}
\vspace{-1.0em}
\end{figure}

\begin{figure}[!t]
  \centering
  \includegraphics[width=\linewidth]{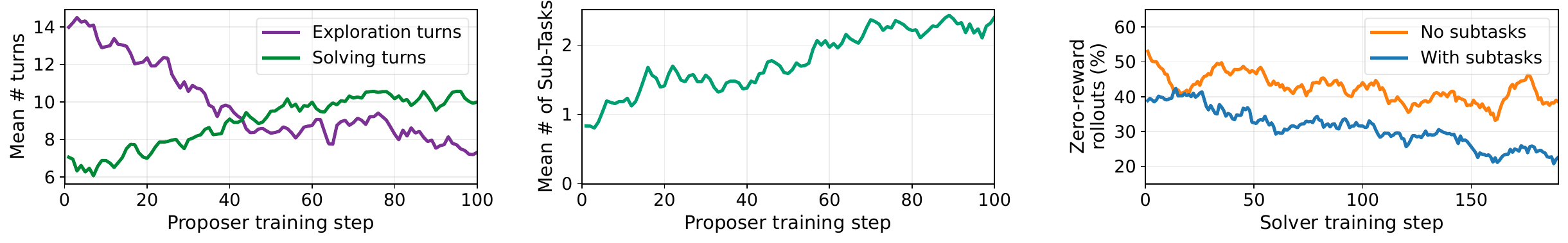}
  \caption{
    Proposer and solver training dynamics on AppWorld.
    \textbf{Left:} The proposer shifts its effort from environment exploration toward executing the task it proposes: exploration turns decrease from $14$ to $7$, while solving turns rise from $7$ to $10$.
    \textbf{Center:} The mean number of declared subtasks increases from around $0.8$ to $2.3$.
    \textbf{Right:} The fraction of zero-reward solver rollouts with subtask supervision remains consistently below the no-subtask ablation, ending near $22\%$ versus $39\%$.
    }
  \label{fig:training-dynamics}
\vspace{-1.5em}
\end{figure}

\subsection{Subtask Supervision Produces Task-Relevant Compositions}

Rewarding subtask declarations could encourage the proposer to insert irrelevant intermediate steps rather than construct meaningful task compositions. However, the left two panels of \cref{fig:analysis2} show that proposer trajectories have similar lengths with and without subtask supervision, while the subtask variant makes fewer API calls on average. The difference lies in the relevance of those interactions to the final task. The right two panels show that subtask supervision increases the fraction of contributing interactions, identified by backward program slicing \citep{weiser1984program}, indicating that the proposer incorporates a larger share of its trajectory into the final task. A complementary analysis finds that dependencies across subtasks increase during training without broader API or application coverage (\cref{app:task-composition-analysis}). \emph{Together, these findings indicate that subtask supervision encourages task-relevant composition rather than simply adding more interactions.}

\subsection{The Proposer Learns to Construct Structured Tasks}

During self-play, the proposer shifts from exploring the environment to executing its proposed tasks (\cref{fig:training-dynamics}, left).
Exploration turns (before $\declaretasktool$) decrease from roughly $14$ to $7$, while solving turns (from $\declaretasktool$ to $\finishtool$) increase from around $7$ to $10$.
Meanwhile, the average number of declared subtasks rises from approximately $0.8$ to $2.3$ (center).
These trends indicate that \emph{the proposer increasingly allocates its interactions to task execution while constructing tasks with more explicit compositional structure}.
Trends in filter acceptance and task-description length are in \cref{app:proposer-dynamics}.\looseness=-1

\subsection{Subtask Rewards Preserve Learning Signal}

We next examine the optimization mechanism developed in \cref{sec:reward-collapse}.
With task-level binary rewards, every unsuccessful rollout receives zero reward regardless of its progress.
\methodname distinguishes partially successful trajectories by rewarding the subtask checkpoints they reach.
In \cref{fig:training-dynamics} (right), the fraction of zero-reward solver rollouts with subtask supervision remains consistently below the no-subtask ablation and ends near $22\%$, compared with $39\%$ without subtasks.
\emph{By rewarding differences in subtask completion, \methodname preserves a learning signal even when every sampled rollout fails to complete the task.}

\begin{figure}[!t]
  \centering
  \includegraphics[width=0.95\linewidth]{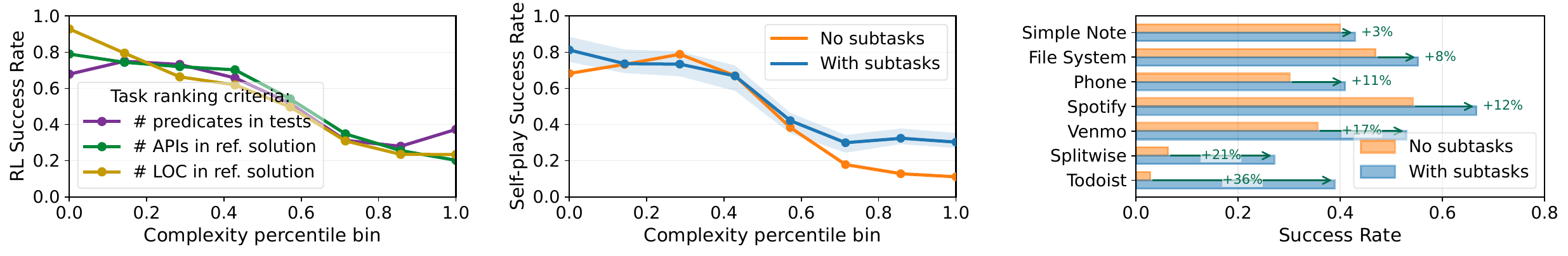}
  \caption{
  \textbf{Left:} On AppWorld, tasks with greater multi-step solution complexity (measured by the number of test-case predicates, API calls, or lines of code in their reference solutions) are substantially harder for the RL agent trained on human-authored tasks.
  \textbf{Center:} Subtask supervision in \methodname provides little consistent benefit on simpler tasks but substantially higher success in the highest-complexity bins.
    \textbf{Right:} Per-app success rates on AppWorld \texttt{test\_normal}, where each task counts toward every app it requires.
    The largest gains occur for Todoist and Splitwise, which are the only apps appearing exclusively in multi-app tasks.
    }
  \label{fig:per-app}
\vspace{-1.0em}
\end{figure}

  \subsection{Benefits Concentrate on Compositional Tasks}

Finally, on AppWorld, success rates for an agent trained with RL on human-authored tasks decrease sharply with multi-step solution complexity (\cref{fig:per-app}, left).
We use three complementary proxies: evaluation predicates count the conditions the final state must satisfy, while API calls and lines of code in reference solutions capture the required operations and computation.
  Comparing self-play with and without subtask supervision (center), the variants perform similarly on simpler tasks, but subtask supervision yields a large and consistent advantage in the highest-complexity bins.
  Its benefits thus concentrate where sparse task-level feedback is expected to be least informative.

  The per-app breakdown provides a complementary view of composition through tasks spanning multiple applications (\cref{fig:per-app}, right).
  The largest gains occur on Todoist and Splitwise, followed by Venmo.
  These applications predominantly appear in multi-app tasks; Todoist and Splitwise appear exclusively in such tasks in this evaluation set.
  These tasks require carrying information and effects across application boundaries, a demanding form of composition.
  Together with the structural analyses, these results suggest that \emph{subtask supervision encourages task-relevant dependencies, with its largest gains on tasks requiring multiple dependent operations}.

\section{Conclusion}
\label{sec:conclusion}
We presented \methodname, which uses environment interaction to jointly generate compositional tasks and fine-grained, verifiable rewards.
  Our theoretical analysis characterizes when subtask rewards preserve a learning signal despite task failure. Experiments show that subtask supervision encourages task-relevant dependencies, with the largest gains on tasks requiring multiple dependent operations.
  Across three benchmarks, \methodname outperforms the evaluated self-play baselines without human-annotated training tasks and improves subsequent human-data RL.
  These findings support using task-generation trajectories as a source of both compositional tasks and intermediate supervision.\looseness=-1

\myparagraph{Generative AI Statement. }
We used generative AI tools to refine conceptual framing and hypotheses, provide feedback on experimental methodology, review mathematical claims and proofs, assist with reference verification, interpret experimental results, and assist with code implementation and review. 
LLMs also generated training tasks and trajectories as part of the self-play method described in \cref{sec:method}. 
We did not use generative AI tools for translation or for cleaning or reformatting datasets. 
We used these tools for literature discovery, manuscript organization, drafting, and readability improvements. The authors derived the proofs manually, conducted their own quantitative and qualitative analyses of experimental results, and independently checked references, code, and literature recommendations. We take responsibility for the final content of this work, including text, claims, and artifacts produced with the assistance of generative AI.

\myparagraph{Ethics Statement. }
Our experiments study agent training in benchmark environments rather than deployment in live services. Nevertheless, improving autonomous agents' ability to execute long-horizon tasks creates potential risks from unintended actions or misuse. The reference-state checks used for training measure task completion, not whether a task is safe or appropriate. Deployment beyond these benchmarks would therefore require additional safeguards, including access controls, restrictions on permitted actions, and human oversight for consequential operations.

\myparagraph{Reproducibility Statement. }
The task-generation procedure, reference-state verification, and reward functions are described in \cref{sec:method}, with the alternating training algorithm in \cref{app:algorithm}. The experimental setup specifies the benchmarks, evaluation protocols, and baseline adaptations (\cref{sec:experimental-setup}); training hyperparameters and hardware are reported in \cref{app:details}, and prompt templates in \cref{app:prompts}. The assumptions and proof of the reward-collapse analysis are provided in \cref{sec:reward-collapse,app:proof}. Additional definitions and procedures for the task-composition analysis appear in \cref{app:task-composition-analysis}.

\myparagraph{Acknowledgements. }
We thank Amir Jalalirad and Gabriele Cesa for insightful discussions and comments during the project.

\bibliography{custom}
\bibliographystyle{iclr2027_conference}

\appendix
\crefalias{section}{appendix}
\crefalias{subsection}{appendix}

\section{Preliminaries}
\label{app:prelim}

\myparagraph{Reinforcement learning with verifiable rewards.}
The general framework of agentic training on long-horizon tasks is reinforcement learning with verifiable rewards (RLVR) \citep{guo2025deepseek}.
We consider an agent, given by a policy $\policy_{\param}$ with parameters $\param$, interacting with a stateful environment $\env$.
Given a task $\task$, the agent produces a trajectory $\traj = (\actionsym_1, \obs_1, \dots, \actionsym_T, \obs_T)$, alternating between actions $\actionsym_t \sim \policy_{\param}(\cdot \mid \task, \actionsym_{<t}, \obs_{<t})$ and observations $\obs_t$ returned by $\env$.
Rather than relying on a learned reward model, RLVR assumes a deterministic verifier that assigns a binary reward $\reward(\task, \traj) \in \{0, 1\}$ to a \emph{completed trajectory}, e.g., by checking the final answer in mathematical reasoning or by executing unit tests in code generation.
The training objective is the expected verifiable reward
\begin{equation}
    \objective_{\textnormal{RLVR}}(\param) = \E_{\task \sim \taskdist}\, \E_{\traj \sim \policy_{\param}(\cdot \mid \task)} \big[ \reward(\task, \traj) \big],
\end{equation}
where $\taskdist$ is a task distribution, typically a human-curated training set.%

\myparagraph{Group relative policy optimization.}
We optimize $\objective_{\textnormal{RLVR}}$ with group relative policy optimization (GRPO) \citep{shao2024deepseekmath}, which forgoes a learned value function and estimates advantages relative to a group of rollouts instead.
For each task $\task$, we sample $\groupsize$ trajectories $\{\traj_i\}_{i=1}^{\groupsize}$ from the previous policy $\oldpolicy$ and compute the group-normalized advantage
\begin{equation}
    \advantage_i = \frac{\reward_i - \mean\big(\{\reward_j\}_{j=1}^{\groupsize}\big)}{\std\big(\{\reward_j\}_{j=1}^{\groupsize}\big)},
    \qquad \reward_i = \reward(\task, \traj_i).
\end{equation}
The policy is updated by maximizing the clipped surrogate objective
\begin{equation}
    \objective_{\textnormal{GRPO}}(\param) = \E_{\task,\, \{\traj_i\} \sim \oldpolicy} \bigg[ \frac{1}{\groupsize} \sum_{i=1}^{\groupsize} \min\Big( \ratio_i \advantage_i,\; \clip\big(\ratio_i, 1-\threshold_1, 1+\threshold_2\big) \advantage_i \Big) \bigg] - \beta\, \mathrm{KL}\big(\policy_{\param} \,\|\, \policy_{\textnormal{ref}}\big),
\end{equation}
where $\ratio_i = \policy_{\param}(\traj_i \mid \task) / \oldpolicy(\traj_i \mid \task)$ is the importance ratio, $\threshold_1, \threshold_2$ are the clipping thresholds, and $\beta$ weights a KL penalty toward a reference policy $\policy_{\textnormal{ref}}$ (\cref{tab:hyperparams}).
In practice, the ratio and clipping are applied per token, and only agent-generated tokens contribute to the objective while environment observations are masked out.

\section{Proof of \texorpdfstring{\cref{prop:zerograd}}{Proposition 1}}
\label{app:proof}

Rollouts in a group are i.i.d.\ draws from $\solverpolicy(\cdot \mid \task)$, so their rewards are i.i.d.\ random variables.
A group of rollouts undergoes reward collapse when all $\groupsize$ rewards are equal.

Under the binary reward, all rewards are equal if and only if all rollouts succeed or all fail, which has probability $p^{\groupsize} + (1-p)^{\groupsize}$ by independence.
As $p \to 0$, $(1-p)^{\groupsize} \to 1$; at $p = 0$ the probability equals $1$.

Conditional on the all-fail event, each reward equals the subgoal fraction and takes values in $\{0, 1/\numsubtasks, \dots, 1\}$ with conditional masses $\tilde{q}_0, \dots, \tilde{q}_{\numsubtasks}$.
For i.i.d.\ discrete variables, the probability that all $\groupsize$ values coincide is $\sum_{j} \tilde{q}_j^{\,\groupsize}$.
If two levels $a \neq b$ have $\tilde{q}_a, \tilde{q}_b > 0$, then every $\tilde{q}_j < 1$, hence $\tilde{q}_j^{\,\groupsize} < \tilde{q}_j$ for every $j$ with $\tilde{q}_j > 0$, and summing gives $\sum_j \tilde{q}_j^{\,\groupsize} < \sum_j \tilde{q}_j = 1$.
\qed

The comparison above is conditional on all sampled rollouts failing. It does not imply that subtask rewards always reduce the unconditional collapse probability: under \cref{eq:solver-reward}, a failed rollout that completes every subtask receives the same reward as a successful rollout.

\myparagraph{Effect of group size.}
  Increasing the group size reduces collapse under binary rewards, but becomes costly as tasks become harder.
  Let
  \[
      C_{\mathrm{bin}}(p,\groupsize)
      = p^{\groupsize} + (1-p)^{\groupsize}
  \]
  denote the binary-reward collapse probability.
  For any target collapse probability $\delta \in (0,1)$, the condition
  $C_{\mathrm{bin}}(p,\groupsize) \leq \delta$ necessarily implies
  \[
      (1-p)^{\groupsize} \leq \delta
      \qquad\Longrightarrow\qquad
      \groupsize \geq
      \frac{\log \delta}{\log(1-p)}.
  \]
  Since $\log(1-p) = -p + O(p^2)$ as $p \to 0$, the required group size grows as
  \[
      \groupsize = \Omega\left(\frac{\log(1/\delta)}{p}\right).
  \]
  Thus, keeping collapse below a fixed threshold requires the number of rollouts per task to grow inversely with the solver's success probability.
  For illustration, if $p=0.01$, the collapse probability is approximately $0.85$ with the group size $\groupsize=16$ used in our experiments and remains approximately $0.53$ with $
  \groupsize=64$.
  Reducing it below $0.05$ requires at least $299$ rollouts per task.

  Subtask rewards can reduce collapse without increasing $\groupsize$.
  Let $\tilde{q}_{\max}=\max_j \tilde{q}_j$.
  Conditional on all rollouts failing,
  \[
      \sum_{j=0}^{\numsubtasks}\tilde{q}_j^{\,\groupsize}
      \leq
      \tilde{q}_{\max}^{\,\groupsize-1}
      \sum_{j=0}^{\numsubtasks}\tilde{q}_j
      =
      \tilde{q}_{\max}^{\,\groupsize-1}.
  \]
  Whenever failed rollouts reach more than one subtask-completion level, $\tilde{q}_{\max}<1$, and the probability of collapse decreases with the group size while already remaining
  strictly below the binary reward's conditional collapse probability of one.
  The magnitude of the reduction depends on how broadly failed rollouts are distributed across subtask-completion levels.

\section{Training Algorithm}
\label{app:algorithm}

\myparagraph{Solver training and co-evolution.}
The two agents are trained in alternating phases.
In the proposer phase, each proposed task is scored by sampling $\numscore = 8$ rollouts from the solver, which is held fixed at its most recent checkpoint; these rollouts serve only to estimate the solve rate $\solverate(\task)$ in the proposer reward (\cref{eq:proposer-reward}).
In the solver phase, the solver draws fresh rollouts on the pool of proposed tasks and is updated with GRPO under the subtask-level reward of \cref{eq:solver-reward}.
Tasks that fail a validity check---malformed outputs or unverifiable completion conditions---are discarded from solver training, and the proposer receives zero reward for them.
The two agents are updated in alternation: as the solver improves, previously challenging tasks become easy, and the proposer is driven to explore further and compose longer subtask sequences, sustaining an automatic curriculum.
The full procedure is given in \cref{alg:selfplay}.
\begin{algorithm}[t]
\caption{Self-play with subtask annotation (\methodname)}
\label{alg:selfplay}
\begin{algorithmic}[1]
\Require environment $\env$, base model, outer loops $N$, proposer/solver training steps $M_{\textnormal{P}}, M_{\textnormal{S}}$, group size $\groupsize$, scoring rollouts $\numscore$, exploration turn budget $T_{\textnormal{P}}$
\State Initialize proposer $\proposerpolicy$ and solver $\solverpolicy$ from the base model
\For{$n = 1, \dots, N$}
    \For{$m = 1, \dots, M_{\textnormal{P}}$} \Comment{proposer phase}
        \State Reset $\env$ to initial state $\state_0$
        \State $\proposerpolicy$ explores $\env$ for up to $T_{\textnormal{P}}$ steps, producing $\proptraj$; a $\declaretasktool$ call ends exploration and states a candidate task; each $\subfinishtool(\subtask_k)$ call snapshots a reference state $\stateref_k$; a final $\finishtool(\task)$ call declares the task $\task$ and snapshots the final reference state $\finalstateref$
        \State Sample $\numscore$ solver rollouts $\{\traj_i\}_{i=1}^{\numscore} \sim \solverpolicy(\cdot \mid \task)$ from $\state_0$ and compute the solve rate $\solverate(\task)$
        \State Compute $\proposerreward(\task,\proptraj)$ (\cref{eq:proposer-reward}) and update $\proposerpolicy$ with GRPO, grouping proposals that start from the same $\state_0$
    \EndFor
    \For{$m = 1, \dots, M_{\textnormal{S}}$} \Comment{solver phase}
        \State Sample a proposed task $\task$ with its initial state $\state_0$, subtasks $\{\subtask_k\}_{k=1}^{\numsubtasks}$, reference states $\{\stateref_k\}_{k=1}^{\numsubtasks}$, and final reference state $\finalstateref$
        \State Sample $\groupsize$ rollouts $\{\traj_i\}_{i=1}^{\groupsize} \sim \solverpolicy(\cdot \mid \task)$, each after resetting $\env$ to $\state_0$; check the final state against $\finalstateref$ and each subtask against $\stateref_k$
        \State Compute $\solverreward(\task, \traj_i)$ (\cref{eq:solver-reward}) and update $\solverpolicy$ with GRPO
    \EndFor
\EndFor
\State \Return solver $\solverpolicy$
\end{algorithmic}
\end{algorithm}

\section{Implementation Details}
\label{app:details}

The proposer and solver are both initialized from Qwen3-4B on ALFWorld and ScienceWorld and from Qwen3-4B-Instruct-2507 or Qwen3-30B-A3B-Instruct-2507 on AppWorld \citep{yang2025qwen3}, and all trained baselines start from the same base model. Both policies are trained with LoRA adapters of rank $32$ and scaling $\alpha = 32$.
For read-only AppWorld subtasks, the expected value is the variable named by \texttt{value\_of} in the proposer prompt (\cref{fig:prompt-appworld-proposer}).
Self-play training runs for four outer loops, each of which first updates the proposer and then the solver.
Prompt templates are given in \cref{app:prompts}.
All training hyperparameters are listed in \cref{tab:hyperparams}.
The reported wall-clock times are per training step on $8\times$H100 GPUs, i.e., roughly $15$ GPU-node hours per ALFWorld run and $29$ per AppWorld run.

\begin{table}[h]\centering\small
\caption{Hyperparameters of \methodname. A single value applies to all environments.}
\label{tab:hyperparams}
\begin{tabular}{@{}lcc@{}}
\toprule
 & ALFWorld, ScienceWorld & AppWorld \\ \midrule
Base model & 4B & \begin{tabular}[c]{@{}c@{}}4B,\\ 30B-A3B\end{tabular} \\
Outer self-play loops $N$ & \multicolumn{2}{c}{$4$} \\
Proposer / solver steps per loop $M_{\textnormal{P}}$, $M_{\textnormal{S}}$ & $25$ / $25$ & $15$ / $50$ \\
Proposer exploration budget $T_{\textnormal{P}}$ (turns) & $30$  &  $40$ \\
Max solver turns & $30$  & $40$ \\
Context length (tokens) & \multicolumn{2}{c}{$40{,}960$} \\ \midrule
GRPO group size $\groupsize$ & \multicolumn{2}{c}{$16$} \\
Solver rollouts per proposed task $\numscore$ & \multicolumn{2}{c}{$8$} \\
Rollouts per training step & \multicolumn{2}{c}{$16$} \\
Batch size & \multicolumn{2}{c}{$16$} \\
Fine-tuning & \multicolumn{2}{c}{LoRA (rank $32$, $\alpha = 32$)} \\
Optimizer & \multicolumn{2}{c}{AdamW} \\
Learning rate & \multicolumn{2}{c}{$10^{-4}$} \\
Clipping thresholds $\threshold_1$ = $\threshold_2$ & \multicolumn{2}{c}{$0.3$} \\
KL penalty & \multicolumn{2}{c}{0.001} \\
Subtask reward weight $\subtaskweight$ & \multicolumn{2}{c}{$0.3$} \\
Diversity reward weight $\diversityweight$ & \multicolumn{2}{c}{$0.3$} \\ \midrule
Hardware & \multicolumn{2}{c}{$8\times$ NVIDIA H100} \\
Wall-clock per proposer step & \multicolumn{2}{c}{$\approx 500$\,s} \\
Wall-clock per solver step & \multicolumn{2}{c}{$\approx 200$\,s} \\
\bottomrule
\end{tabular}
\end{table}

\section{Additional Results}

\subsection{Training Curves}
\label{app:curves}
\begin{figure}[t]
  \centering
  \includegraphics[width=0.80\linewidth]{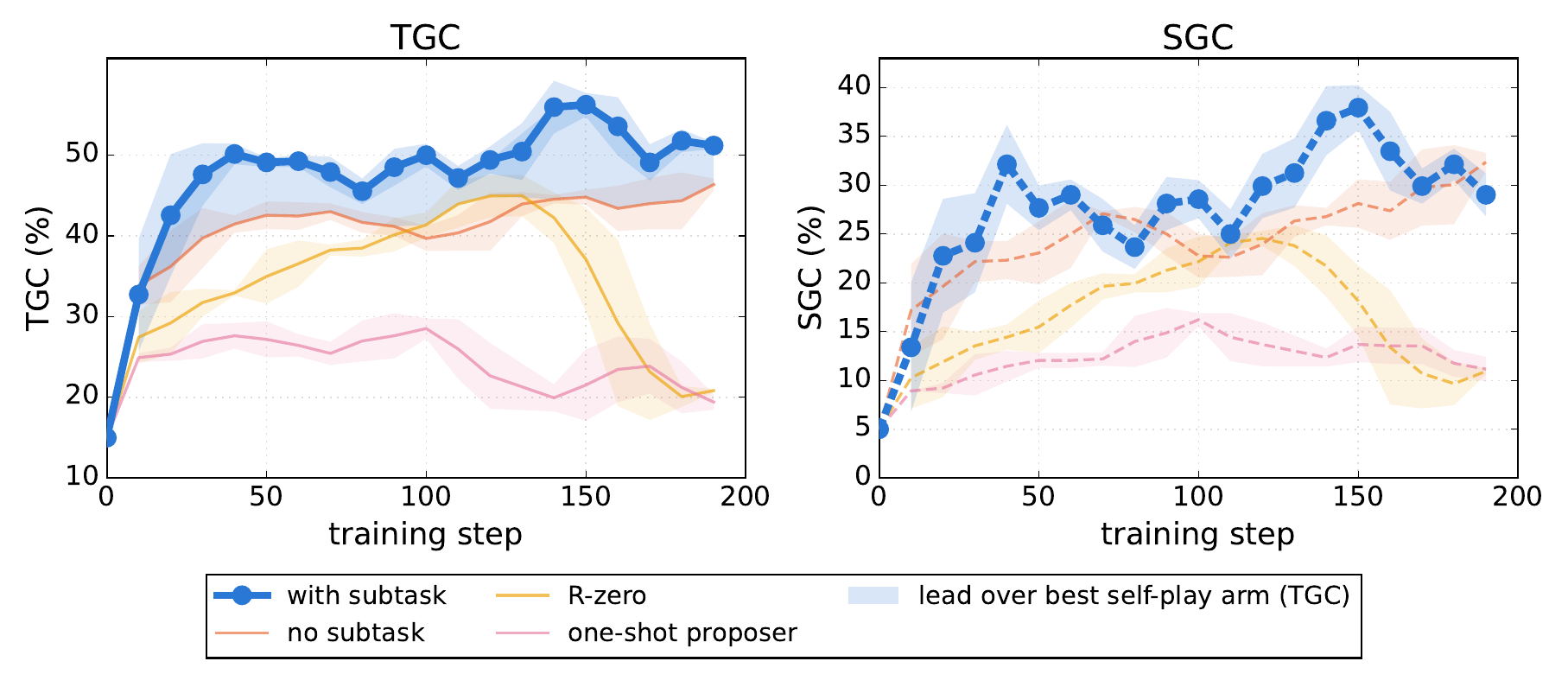}
  \caption{
  Learning curves on AppWorld (Qwen3-30B-A3B-Instruct-2507) (left: TGC, right: SGC; shaded bands: standard error).
  Curves evaluate \emph{one single rollout per task} at each checkpoint, so endpoints are not directly comparable to the reported averaged metric in \cref{tab:main}.
  Without any human-annotated task, the full method (\emph{with subtask}) keeps improving throughout training, whereas removing the subtask reward (\emph{no subtask}) plateaus earlier and converges to lower scores.
  \emph{R-Zero} and \emph{one-shot proposer} verify tasks against solver-rollout consensus states instead of proposer snapshots, with interactively and one-shot generated tasks, respectively.%
  }
  \label{fig:appworld-curve}
\end{figure}

\Cref{fig:appworld-curve} examines these differences over training with the 30B AppWorld agent.
The full method keeps improving in both TGC and SGC throughout the training, while the interactive variant without subtasks remains near its performance level after the first 60 steps.
The one-shot proposer improves initially but loses much of that gain later.
Despite fluctuations during training, the full method's late improvement suggests that subtask supervision supports progress beyond the initial gains from self-play.
We use these checkpoint evaluations to examine training dynamics and \cref{tab:main} for the final performance comparisons.
Qualitative examples of generated tasks are provided in \cref{app:qual}.

\subsection{Additional Analysis of Generated Task Composition}
\label{app:task-composition-analysis}

On the proposer's own AppWorld exploration rollouts, tasks become more compositional over training.
We compute three structural diagnostics directly from the executed Python steps, recorded environment deltas, and proposer-declared subtask boundaries:
\begin{enumerate}[leftmargin=*,nosep]
      \item \emph{Environment-grounded}: a declared subtask is counted when it is supported by a task-relevant AppWorld API and records either a nonempty environment delta or a value obtained from that API. For read-only subtasks, backward data-flow tracing verifies that the recorded value originates from an environment API rather than from a locally introduced constant. Setup operations such as authentication, documentation lookup, and supervisor access are excluded.
      \item \emph{Verified}: a declared checkpoint is counted when it records an environment state change and the final task adds at least one further state change beyond that checkpoint. A read-only checkpoint can instead qualify through a captured value whose provenance traces to a task-relevant API. Checkpoints must occur at distinct, increasing execution steps.
      \item \emph{Dependency-chained}: we parse each executed Python step into the variables it binds and uses. A subtask boundary is counted when a variable returned by a task-relevant API before the boundary is used in a later step. This captures information flow across components, such as retrieving an address from Contacts and subsequently using it to place an order.
\end{enumerate}
From the first to the last $10$-step training block, the mean number of dependency-chained segments per proposal increases from $0.57$ to $1.13$, environment-grounded subgoals from $0.55$ to $1.22$, and verified segments from $0.29$ to $0.77$; the fraction of well-posed proposals rises from $56\%$ to $77\%$.
This growth is not explained by broader exploration: distinct APIs per task decrease from $5.9$ to $5.2$, and the fraction of multi-app tasks decreases from $27\%$ to $18\%$.
Instead, the proposer increasingly combines operations within a task into deeper dependency structures.
Consistent with this interpretation, the proposer declares subgoals for a growing fraction of tasks ($53\%$ to $74\%$), and the mean number of declared subgoals rises from $0.89$ to $1.19$.

A control proposer trained without the subtask reward exhibits a comparable increase in well-posed proposals ($57\%$ to $79\%$) and execution success ($85\%$ to $93\%$), suggesting that the compositional trend is not merely a consequence of proposals becoming better formed.
Its exploration also narrows more rapidly: effective APIs per task fall by $26\%$, compared with $18\%$ with subtask supervision, while the number of applications that provide values to later steps falls by $52\%$, compared with $30\%$.
These measurements suggest that subtask supervision encourages dependency structure while slowing the loss of exploratory breadth during proposer training.

\subsection{Additional Proposer Training Dynamics}
\label{app:proposer-dynamics}

\begin{figure}[t]
  \centering
  \includegraphics[width=0.8\linewidth]{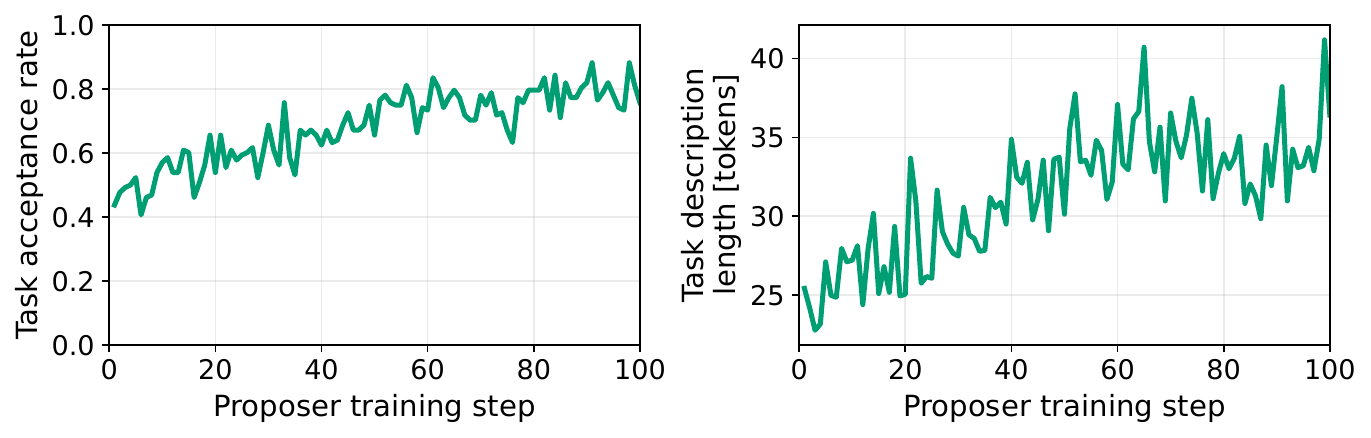}
  \caption{
    Additional proposer training dynamics on AppWorld.
    \textbf{Left:} The fraction of proposed tasks accepted for solver training increases from roughly $45\%$ early in training to around $80\%$ in later stages.
    \textbf{Right:} Task descriptions become longer over training, rising from roughly $25$ tokens to around $35$, with substantial variation between steps.
    }
  \label{fig:proposer-acceptance-description}
\end{figure}

\myparagraph{Task acceptance.}
The acceptance rate is the fraction of proposals that pass the rule-based filter described in \cref{sec:solver} and can be used for solver training.
The filter requires a final task description, a non-redundant subtask checklist, and that applications or objects mentioned in the description also occur in the proposer trajectory.
Under these criteria, acceptance rises from roughly $45\%$ to around $80\%$ over training (\cref{fig:proposer-acceptance-description}, left).
This trend indicates that the proposer increasingly generates tasks that meet the requirements for inclusion in the solver's training distribution.

\myparagraph{Task-description length.}
Descriptions also become longer, increasing from roughly $25$ tokens early in training to around $35$ in later stages (\cref{fig:proposer-acceptance-description}, right).
This trend accompanies the increase in declared subtasks shown in \cref{fig:training-dynamics}.
Longer descriptions are consistent with specifying more task components, but token count alone does not establish compositionality or instruction quality.
The evidence for task relevance and downstream utility comes from the analyses in \cref{sec:empirical-analysis}.

\section{Qualitative Examples}
\label{app:qual}

\Cref{fig:qual-phase1,fig:qual-phase3} show truncated proposer explorations from the first and third self-play phase on AppWorld, in the todoist and spotify apps respectively, and \cref{fig:qual-alf-phase1,fig:qual-alf-phase3} show the same for ALFWorld.
The subtask-declaration tool, written $\subfinishtool$ in the main text, is exposed to the model under the name \texttt{sub\_finish}, which is what the prompts and the 4B listings show; the 30B listings write it as \texttt{subtask} and the task-declaration tool as \texttt{declare\_task}.
In all listings, the proposer's tool calls and reasoning are shown on a white background, and environment responses are \colorbox{envbg}{shaded}.
In phase 1, the proposer inspects the API documentation, logs in, lists projects and tasks, and creates a single task, retrying once after an invalid-parameter error; it declares no subtasks, and the resulting task requires a single state change.
In phase 3, the proposer declares three subtasks along the way (logging in, surveying the user's liked songs and playlists, and creating a playlist), creates the playlist, adds four songs to it, and finishes with a task that names the playlist and all four songs to be added.
On ALFWorld, the phase-1 proposer picks up a spoon, fails to place it in a cabinet (``Nothing happens''), succeeds with a drawer, and finishes with a single pick-and-place task; the phase-3 proposer retrieves a plate, cools it in the fridge, declares the cooled plate as a subtask, and returns it to the cabinet, composing two operations into a \emph{cool-and-place} task with one intermediate checkpoint.

\Cref{fig:qual-30b-spotify,fig:qual-30b-amazon} show two explorations by the Qwen3-30B-A3B-Instruct-2507 proposer on AppWorld. In these listings the subtask-declaration tool appears as \texttt{subtask}, and each call is placed directly after the step it records. 

\begin{figure}[H]
\begin{minted}{text}
# Proposer exploration, phase 1 (AppWorld / todoist).  Tool calls: 8, subtask: 0.
>>> print(apis.api_docs.show_app_descriptions())
>>> print(apis.api_docs.show_api_descriptions(app_name='todoist'))
>>> supervisor_passwords = apis.supervisor.show_account_passwords()
>>> todoist_login = apis.todoist.login(username="ch-burch@gmail.com", password=...)
>>> projects = [item for page_index in range(0, 10)
...     for item in apis.todoist.show_projects(access_token=..., page_index=page_index)]
>>> inbox_tasks = apis.todoist.show_tasks(access_token=..., project_id=32)  # Inbox project
>>> new_task = apis.todoist.create_task(access_token=..., project_id=32,
...     title="Schedule a weekly mental health check-in", priority="medium",
...     due_date=None, assignee_email=None)
    Exception: Unexpected parameter 'assignee_email' passed to the create_task API ...
>>> new_task = apis.todoist.create_task(access_token=..., project_id=32,
...     title="Schedule a weekly mental health check-in", priority="medium", due_date=None)
    {"message": "Task created.", "task_id": 6334}
>>> finish("Create a new task titled 'Schedule a weekly mental health check-in' in the
...     Inbox project of my todoist account with medium priority and no due date.")
    Exploration finished.
\end{minted}
\caption{Proposer exploration in an early self-play phase (phase 1) on AppWorld (todoist) with Qwen3-4B-Instruct-2507. The proposer inspects the API documentation, logs in, lists projects and tasks, and creates a single task, retrying once after an invalid-parameter error; it declares no subtasks, and the final task requires one state change. Shaded lines are environment responses.}
\label{fig:qual-phase1}
\end{figure}

\begin{figure}[H]
\begin{minted}{text}
# Proposer exploration, phase 3 (AppWorld / spotify).  Tool calls: 11, subtask: 3.
>>> print(apis.api_docs.show_api_descriptions(app_name='spotify'))
>>> spotify_login = apis.spotify.login(username="kathrynmaldonado@gmail.com", password=...)
>>> subtask("Login to Spotify using the supervisor's credentials and obtain an access
...             token.")
    Subtask recorded.
>>> liked_songs = [item for page_index in range(0, 10)
...     for item in apis.spotify.show_liked_songs(access_token=...)]
    [{"song_id": 298, "title": "When Dawn Breaks Without You", ...}, {"song_id": 165, ...}]
>>> playlists = [item for page_index in range(0, 10)
...     for item in apis.spotify.show_playlist_library(access_token=...)]
    [{"playlist_id": 526, "title": "Sunset Chill: Songs for Golden Hours", ...}, ...]
>>> subtask("Explore the user's liked songs and playlists to identify musical preferences
...             for creating a meaningful playlist.")
    Subtask recorded.
>>> apis.spotify.create_playlist(title="Evening Reflections: Melancholic Melodies",
...     access_token=...)
    {"message": "Playlist created.", "playlist_id": 654}
>>> subtask("Create a new playlist titled \"Evening Reflections: Melancholic Melodies\" and
...             add relevant songs based on the user's liked songs and emotional themes.")
    Subtask recorded.
>>> apis.spotify.add_song_to_playlist(song_id=298, playlist_id=654, access_token=...)
>>> apis.spotify.add_song_to_playlist(song_id=165, playlist_id=654, access_token=...)
...     # likewise for song_id=183 and song_id=9
    Execution successful.
>>> finish("Create a new playlist titled 'Evening Reflections: Melancholic Melodies' and add
...     the songs 'When Dawn Breaks Without You', 'Hearts on Fire', 'In the Tunnels of Despair',
...     and 'When Fate Becomes a Foe' to it.")
\end{minted}
\caption{Proposer exploration in a late self-play phase (phase 3) on AppWorld (spotify) with Qwen3-4B-Instruct-2507. The proposer declares three subtasks: after logging in, after surveying the user's liked songs and playlist library, and after creating a new playlist (declared before the songs are added, so its reference state is the empty playlist). The login subtask is setup that the prompt asks the proposer not to declare, and the validity filter of \cref{sec:solver} does not remove it. The proposer then adds four liked songs to the playlist and finishes with a task that specifies the playlist title and all four songs. Shaded lines are environment responses.}
\label{fig:qual-phase3}
\end{figure}

\begin{figure}[H]
\begin{minted}{text}
# Proposer exploration, phase 1 (ALFWorld).  Actions: 9, subtask: 0.  Shown as verb(args).
>>> go_to("countertop 1")
    On the countertop 1, you see a bowl 2, a bread 1, a dishsponge 2, a kettle 2, a knife 3, ...
>>> take("spoon 1", from="countertop 1")
    You pick up the spoon 1 from the countertop 1.
>>> go_to("cabinet 1")
    You arrive at cabinet 1. The cabinet 1 is closed.
>>> open("cabinet 1")
    You open the cabinet 1. The cabinet 1 is open. In it, you see a kettle 1.
>>> move("spoon 1", to="cabinet 1")
    Nothing happens.
# proposer: I tried to move the spoon 1 to the cabinet, but it didn't work. ... Let me try a
#           different approach.
>>> go_to("drawer 1")
    You arrive at drawer 1. The drawer 1 is closed.
>>> open("drawer 1")
    You open the drawer 1. The drawer 1 is open. In it, you see a fork 1.
>>> move("spoon 1", to="drawer 1")
    You move the spoon 1 to the drawer 1.
>>> finish("Put a spoon in the drawer.")
    Exploration finished.
\end{minted}
\caption{Proposer exploration in an early self-play phase (phase 1) on ALFWorld with Qwen3-4B. The proposer picks up a spoon from the countertop, fails to place it in a cabinet, recovers by opening a drawer and placing it there, and finishes with a single pick-and-place task. Shaded lines are environment responses.}
\label{fig:qual-alf-phase1}
\end{figure}

\begin{figure}[H]
\begin{minted}{text}
# Proposer exploration, phase 3 (ALFWorld).  Actions: 8, subtask: 1.  Shown as verb(args).
>>> go_to("cabinet 1")
    You arrive at cabinet 1. The cabinet 1 is closed.
>>> open("cabinet 1")
    You open the cabinet 1. The cabinet 1 is open. In it, you see a plate 1.
>>> take("plate 1", from="cabinet 1")
    You pick up the plate 1 from the cabinet 1.
# proposer: I now have the plate. I can move it to another location. Let me go to the fridge to
#           cool the plate.
>>> go_to("fridge 1")
    You arrive at fridge 1. The fridge 1 is closed.
>>> cool("plate 1", with="fridge 1")
    You cool the plate 1 using the fridge 1.
>>> subtask("Cool a plate.")
    Subtask recorded.
>>> go_to("cabinet 1")
    You arrive at cabinet 1. The cabinet 1 is open. In it, you see nothing.
>>> move("plate 1", to="cabinet 1")
    You move the plate 1 to the cabinet 1.
>>> finish("Cool a plate and put it in a cabinet.")
    Exploration finished.
\end{minted}
\caption{Proposer exploration in a late self-play phase (phase 3) on ALFWorld with Qwen3-4B. The proposer retrieves a plate from a cabinet, cools it using the fridge and declares this as a subtask, then returns the plate to the cabinet, finishing with a task that composes two operations (\emph{cool} and \emph{put}). Shaded lines are environment responses.}
\label{fig:qual-alf-phase3}
\end{figure}

\begin{figure}[H]
\begin{minted}{text}
# Proposer exploration, phase 1 (AppWorld / spotify, 30B). Tool calls: 18, subtask: 2.
>>> spotify_access_token = apis.spotify.login(username="jes.mill@gmail.com",
...     password=...)["access_token"]
>>> song_library = [item for page_index in range(0, 10)
...     for item in apis.spotify.show_song_library(page_index=page_index, access_token=...)]
    Collected 18 songs from library.
>>> playlist_library = [item for page_index in range(0, 10)
...     for item in apis.spotify.show_playlist_library(page_index=page_index, access_token=...)]
    Collected 5 playlists.
>>> for playlist in playlist_library: playlist_song_ids |= set(playlist["song_ids"])
>>> unplanned_song_ids = {song["song_id"] for song in song_library} - playlist_song_ids
    Number of songs not in any playlist: 16
>>> declare_task("Create a new Spotify playlist titled \"Missing in Action\" and add every song
...     from my library that is not in any of my playlists.")
>>> playlist_id = apis.spotify.create_playlist(title="Missing in Action", is_public=False,
...     access_token=...)["playlist_id"]
    Created playlist with ID: 654
>>> subtask("Create a new Spotify playlist titled \"Missing in Action\".")
    Subtask recorded.
>>> for song_id in unplanned_song_ids:
...     apis.spotify.add_songs_to_playlist(playlist_id=playlist_id, song_ids=[song_id], ...)
    Exception: No API named 'add_songs_to_playlist' found in the spotify app.
>>> print(apis.api_docs.show_api_descriptions(app_name='spotify'))
>>> for song_id in unplanned_song_ids:
...     apis.spotify.add_song_to_playlist(playlist_id=playlist_id, song_id=song_id, ...)
>>> subtask("Add every song from my library that is not in any playlist to the
...             \"Missing in Action\" playlist.")
    Subtask recorded.
>>> ...  # read the playlist back and compare with unplanned_song_ids
    Match: True
>>> finish(...)
    Exploration finished.
\end{minted}
\caption{Proposer exploration in the first self-play phase on AppWorld (spotify) with Qwen3-30B-A3B-Instruct-2507. The proposer computes the library songs that are in no playlist, declares a task conditioned on this set, creates the playlist and adds the songs, declaring one subtask for each step, and verifies the result before finishing. Shaded lines are environment responses.}
\label{fig:qual-30b-spotify}
\end{figure}

\begin{figure}[H]
\begin{minted}{text}
# Proposer exploration, phase 1 (AppWorld / amazon+todoist, 30B).
# Tool calls: 19, subtask: 2.
>>> amazon_access_token = apis.amazon.login(username='matthew.blac@gmail.com',
...     password=...)['access_token']
>>> page = apis.amazon.show_order_history(access_token=..., page_index=page_index)
    Exception: No API named 'show_order_history' found in the amazon app.
>>> print(apis.api_docs.show_api_descriptions(app_name='amazon'))
>>> page = apis.amazon.show_orders(access_token=..., page_index=page_index)
    Total delivered orders: 0
>>> cart_items = apis.amazon.show_cart(access_token=...)
    {"base_cost": 8390.7, ..., "cart_items": [{"product_name": "Picnic Time Country ...
>>> wish_list_items = apis.amazon.show_wish_list(access_token=...)
    Total items in wish list: 1
>>> for cart_item in cart_items["cart_items"]:
...     apis.amazon.delete_product_from_cart(access_token=...,
...         product_id=cart_item["product_id"])
    All items removed from Amazon cart.
>>> subtask("Remove all items from the Amazon cart.")
    Subtask recorded.
>>> task = apis.todoist.create_task(access_token=...,
...     content="Review high-cost Amazon purchases", due_date="2023-05-25")
    Exception: Unexpected parameter 'content' passed to create_task. Allowed parameters
    are: ['project_id', 'title', ...]
>>> task = apis.todoist.create_task(access_token=..., title="Review high-cost Amazon ...", ...)
    Response status code is 422: project_id: Input should be a valid integer ...
>>> projects = apis.todoist.show_projects(access_token=...)
    [..., {"name": "Inbox", "is_inbox": true, "project_id": 41, ...}, ...]
>>> task = apis.todoist.create_task(access_token=..., title="Review high-cost Amazon purchases",
...     due_date="2023-05-25", project_id=41)
    Todoist task created in Inbox.
>>> subtask("Create a Todoist task titled 'Review high-cost Amazon purchases' with a due
...             date of 2023-05-25 in the Inbox project.")
    Subtask recorded.
>>> finish("Remove all items from my Amazon cart and create a Todoist task titled
...     'Review high-cost Amazon purchases' due on 2023-05-25 to follow up on the
...     cleared items.")
    Exploration finished.
\end{minted}
\caption{Proposer exploration in the first self-play phase on AppWorld (amazon and todoist) with Qwen3-30B-A3B-Instruct-2507. The proposer clears the Amazon cart and creates a follow-up Todoist task, recovering from three API errors along the way, and declares one subtask per app before finishing with a cross-app task. Shaded lines are environment responses.}
\label{fig:qual-30b-amazon}
\end{figure}

\section{Prompt Templates}
\label{app:prompts}

\Crefrange{fig:prompt-alfworld-proposer}{fig:prompt-appworld-solver} show the prompt templates used in our experiments: the proposer and solver prompts for ALFWorld, ScienceWorld, and AppWorld.
The proposer prompts instruct the model to explore without a given task and expose the $\subfinishtool$ and $\finishtool$ tool calls that declare subtasks and the final task. The AppWorld solver prompt follows the AppWorld benchmark's standard prompt, with its worked example omitted here.
Details are omitted for space; ``\ldots'' marks omitted content.

\begin{figure}[H]
\begin{minted}{text}
system:
  You are an AI agent exploring ALFWorld, a text-based household environment.
  You are NOT given a task.
  Your job is to EXPLORE this scene by interacting with it, and DISCOVER an interesting task
      that could be done here.
  Take a sequence of actions that changes the state of the world ... Aim for about 10 to 20
      actions.
  IMPORTANT: respond in this format.
  - On each exploration step: the <think></think> tags stay empty; write a brief plain-text
      reasoning AFTER them about what interesting, state-changing action to take next, then
      output your action in <tool_call></tool_call> tags.
  Common actions. Each example shows only the `arguments` object -- always emit it wrapped as
      `{"name":"alfworld_tool","arguments":{...}}` inside <tool_call></tool_call> tags:
    - Go To: Move to a location. `{"action_type": "go_to", "receptacle_name": "table 1"}`
    - Take: Pick up an object. Example: `{"action_type": "take", "object_name": "apple 2",
        "receptacle_name": "counter 1"}`
    ... (further actions: look, inventory, open, close, move, examine, use, heat, clean, cool,
        slice)
  WHEN YOU ARE DONE: once you have accomplished something interesting and state-changing, END
      by emitting a single "finish" tool call stating that task:
      `{"name":"alfworld_tool","arguments": {"action_type": "finish", "task_description": "<one
          concise sentence>"}}`
  Write the task_description as an imperative instruction, never as a report: "Put a pen in the
      garbagecan.", never "I put a pen in the garbagecan." ...
  MULTI-STEP TASKS: if the task you are building has more than one sub-step (e.g. FIRST heat an
      object, THEN place it somewhere), call "sub_finish" right after EACH completed sub-step
      to declare it, and reserve "finish" for the very last sub-step:
      `{"name":"alfworld_tool","arguments": {"action_type": "sub_finish", "task_description":
          "<that sub-step, as an imperative instruction>"}}`
  sub_finish records the sub-step WITHOUT ending the episode -- you keep acting afterward.
      Example: heat an apple -> sub_finish("Heat an apple.") -> put it on the countertop ->
      finish("Heat some apple and put it on the countertop.").
user:
  {current_observation}
  Explore this scene. Take one action per turn. When you have reached an interesting
      state-changing result, stop and call `finish` with a `task_description` stating that task
      as an imperative instruction ...
\end{minted}
\caption{ALFWorld proposer prompt template. Details omitted for space.}
\label{fig:prompt-alfworld-proposer}
\end{figure}

\begin{figure}[H]
\begin{minted}{text}
system:
  You are an AI assistant playing ALFWorld, a text-based household environment.
  Your goal is to complete household tasks by interacting with objects.
  IMPORTANT: You MUST respond in the following format:
  1. The <think></think> tags stay empty; write your brief plain-text plan AFTER them
  2. Then, output your action in <tool_call></tool_call> tags
  Common actions. Each example shows only the `arguments` object -- always emit it wrapped as
      `{"name":"alfworld_tool","arguments":{...}}` inside <tool_call></tool_call> tags:
    - Go To: Move to a location. `{"action_type": "go_to", "receptacle_name": "table 12"}`
    - Take: Pick up an object. Example: `{"action_type": "take", "object_name": "apple 2",
        "receptacle_name": "counter 1"}`
    - Heat: Heat an object. Example: `{"action_type": "heat", "object_name": "potato 2",
        "receptacle_name": "microwave 1"}`
    ... (further actions: look, inventory, open, close, move, examine, use, clean, cool, slice)
user:
  # Example Omitted
  # Your Task
  {current_observation}
  Your task is: {task_description}
  Explore the environment and complete the task.
\end{minted}
\caption{ALFWorld solver prompt template. Details omitted for space.}
\label{fig:prompt-alfworld-solver}
\end{figure}

\begin{figure}[H]
\begin{minted}{text}
system:
  You are an agent exploring ScienceWorld, a text-based simulator of a house where you can do
      science: move between rooms, open and close containers, pick things up and move them,
      activate and deactivate devices ..., heat, cool, melt, boil or freeze substances, mix
      ingredients, plant seeds, and connect electrical components.
  You are NOT given a task.
  Your job is to EXPLORE this scene and DISCOVER an interesting task that could be done here, by
      actually doing it. ... Do something of that kind: a sequence of actions that CHANGES
      THE STATE OF THE WORLD, usually across more than one room. Aim for about 8 to 20 actions ...
  Call the `scienceworld_tool` tool exactly once per turn with a single `action` string. ...
  FOCUS FIRST: every task in this house is about one main object ... `focus on <object>` BEFORE
      you change it, and begin your task_description with "First, focus on the <object>. Then
      <what to do>." ...
  ... (find-task rules omitted)
  WHEN YOU ARE DONE: once you have changed the world in an interesting way, END with a single
      `finish` call stating that task:
      action: "finish", task_description: "<one concise sentence>"
  Write the task_description as an imperative instruction, never as a report ... It must name
      the objects, containers and end state you actually reached ...
  MULTI-STEP TASKS: if the task you are building has more than one sub-step (e.g. FIRST move a
      substance into a container, THEN heat it until it boils), call `sub_finish` right after
      EACH completed sub-step to declare it, and reserve `finish` for the very last sub-step:
      action: "sub_finish", task_description: "<that sub-step, as an imperative instruction>"
  sub_finish records the sub-step WITHOUT ending the episode -- you keep acting afterward. ...
      declare each sub-step ONCE, right after it is done ...
user:
  # Scene
  {current_observation}
  Explore this scene. Take one action per turn. When you have changed the world in an
      interesting way, stop and call `finish` with a `task_description` stating that task as an
      imperative instruction, never as a report of what you did.
\end{minted}
\caption{ScienceWorld proposer prompt template. Details omitted for space.}
\label{fig:prompt-scienceworld-proposer}
\end{figure}

\begin{figure}[H]
\begin{minted}{text}
system:
  You are an agent completing persistent household-science tasks in the ScienceWorld
      simulator. Each task gives you a goal to accomplish by moving around a house-like
      environment and interacting with objects ...
  Call the `scienceworld_tool` tool exactly once per turn with a single `action` string.
      Prefer one of the valid actions listed in your latest observation ... Read the
      observation returned after each action before deciding your next move, and keep working
      until the task is complete.
user:
  Task: {task_description}
  {formatted_observation}
\end{minted}
\caption{ScienceWorld solver prompt template. Details omitted for space.}
\label{fig:prompt-scienceworld-solver}
\end{figure}

\begin{figure}[H]
\begin{minted}{text}
system:
  You are an AI assistant exploring the apps connected to a supervisor's accounts (spotify,
      venmo, gmail, amazon, phone, ... and others).
  You are NOT given a task.
  Your job is to SURVEY these apps, DECIDE on a task a supervisor might plausibly ask for,
      and carry it out. ... Take a sequence of
      API calls that CHANGES the supervisor's state ... Aim for a short, coherent task (about 5
      to 15 code steps including the setup).
  You interact through a Python REPL: on each step you write ONE small chunk of Python code
      inside a tool call ... Call `apis.<app>.<api>(...)` to use an app's API.
  ... (API discovery, credentials, response format, coding rules, task-condition rules,
      question-answering tasks, and no-narration rules omitted)
  WHEN YOU ARE DONE: once you have accomplished something interesting and state-changing, END
      by emitting a single `finish` action that STATES that task:
      {"name": "appworld_tool", "arguments": {"action_type": "finish", "task_description":
          "<one concise sentence>"}}
  WRITE IT AS AN IMPERATIVE REQUEST, NEVER AS A REPORT. ...
  MULTI-STEP TASKS: A TASK SPANNING TWO APPS IS THE MOST VALUABLE KIND -- read a value in one
      app, then act on it in another. Declare each sub-step as you finish it with `sub_finish`,
      reserving `finish` for the LAST one:
      {"name": "appworld_tool", "arguments": {"action_type": "sub_finish", "task_description":
          "<that sub-step, as an imperative request>", "value_of": "<the name of variable that
          holds the expected value if the sub-step is about finding information>"}}
  It records the sub-step WITHOUT ending the episode. ... What is NEVER a sub-step is setup
      every task needs -- logging in, reading API docs, fetching your own profile ...
user:
  {current_observation}
  Explore these apps. Take ONE action per turn -- normally an `appworld_tool` call with a
      `code` argument, plus `sub_finish` / `finish` when the moment calls for it. ...
      When you have reached an interesting, state-changing result, stop and call `finish`
      with a `task_description` stating that task as an imperative request, never as a
      report of what you did.
\end{minted}
\caption{AppWorld proposer prompt template. Details omitted for space.}
\label{fig:prompt-appworld-proposer}
\end{figure}

\begin{figure}[H]
\begin{minted}{text}
SYSTEM:
  When you receive a tool call response, use the output to format an answer to the original
      user question.
  You are a helpful assistant with tool calling capabilities.
USER:
  I am your supervisor, and you are an AI Assistant whose job is to complete my day-to-day
      tasks fully autonomously.
  To do this, you will need to interact with app(s) (e.g., spotify, venmo etc) using their
      associated APIs on my behalf. ...
  Here are three key APIs that you need to know to get more information
    print(apis.api_docs.show_app_descriptions())
    print(apis.api_docs.show_api_descriptions(app_name='spotify'))
    print(apis.api_docs.show_api_doc(app_name='spotify', api_name='login'))
  Each code execution will produce an output that you can use in subsequent calls. ...
  My name is: {{ main_user.first_name }} {{ main_user.last_name }}. My personal email is {{
      main_user.email }} and phone number is {{ main_user.phone_number }}.
  Task: How many playlists do I have in my Spotify playlist library?
... [worked example omitted: alternating ASSISTANT code chunks and USER outputs] ...
USER:
  **Key instructions**:
  A. General instructions:
  - Act fully on your own. You must make all decisions yourself and never ask me or anyone else
      to confirm or clarify. ...
  ... (further general, app-specific, and code-operation instructions omitted)
  D. Task-completion instructions:
  You must call the `apis.supervisor.complete_task` API after completing the task.
  - If an answer is needed, e.g., for "How many songs are in the Spotify queue?", call it with
      the appropriate answer argument value.
  - The task is doable, but if you cannot find a way, you can call it with status="fail" to
      exit with failure.
USER:
  Using these APIs, now generate code to solve the actual task:
  Today's date is: {{ date }}
  My name is: {{ main_user.first_name }} {{ main_user.last_name }}. My personal email is {{
      main_user.email }} and phone number is {{ main_user.phone_number }}.
  Task: {{ instruction }}
\end{minted}
\caption{AppWorld solver prompt template. Details omitted for space.}
\label{fig:prompt-appworld-solver}
\end{figure}

\section{Limitations}
\label{app:limitations}
Our approach has limitations that suggest directions for future work.
Snapshot-based verification requires an environment that can be reset and whose state can be recorded and compared, an assumption that holds in simulators but not, e.g., on the live web.
Extending reference-state matching to partially observable or irreversible environments is an open problem.
The proposer prompts (\cref{app:prompts}) carry inductive biases that may make the generated task distribution differ from real-world tasks.
The control run of \cref{sec:empirical-analysis}, trained with the same prompts but without the subtask reward, helps separate the contribution of the subtask reward from that of the prompts. 
We believe grounding self-generated tasks in executed interaction, and reusing that interaction as process supervision, will remain a useful recipe as self-evolving agents move to richer environments.

\end{document}

%% file: math_commands.tex
\usepackage{amsmath,amsfonts,bm}

\def\eqref#1{equation~\ref{#1}}

\def\1{\bm{1}}

\DeclareMathAlphabet{\mathsfit}{\encodingdefault}{\sfdefault}{m}{sl}
\SetMathAlphabet{\mathsfit}{bold}{\encodingdefault}{\sfdefault}{bx}{n}

\def\gS{{\mathcal{S}}}

\newcommand{\E}{\mathbb{E}}

%% file: macro.tex
\newcommand{\defn}[1]{\textbf{#1}}
\newcommand{\myparagraph}[1]{\noindent\textbf{#1}}
\newcommand{\symbolthanks}[2]{%
    \setcounter{footnote}{\numexpr#1-1\relax}%
    \thanks{#2}%
}

\providecommand{\shortcite}[1]{}

\newcommand{\meanstd}[2]{{#1 {\tiny $\pm$ #2}}}

\newcommand{\policy}{{\pi}}
\newcommand{\proposerpolicy}{{\policy_{\textnormal{P}}}}
\newcommand{\solverpolicy}{{\policy_{\textnormal{S}}}}

\newcommand{\param}{{\mathbf{\theta}}}

\newcommand{\advantage}{{A}}
\newcommand{\reward}{{R}}
\newcommand{\std}{{\textnormal{std}}}
\newcommand{\mean}{{\textnormal{mean}}}
\newcommand{\clip}{{\textnormal{clip}}}

\newcommand{\objective}{{\mathcal{L}}}
\newcommand{\threshold}{{\varepsilon}}

\newcommand{\task}{{x}}
\newcommand{\taskdist}{{\mathcal{D}}}
\newcommand{\traj}{{\tau}}
\newcommand{\env}{{\mathcal{E}}}
\newcommand{\actionsym}{{a}}
\newcommand{\obs}{{o}}
\newcommand{\oldpolicy}{{\policy_{\textnormal{old}}}}
\newcommand{\ratio}{{\rho}}
\newcommand{\groupsize}{{G}}
\newcommand{\numscore}{{\groupsize_{\textnormal{P}}}} %

\newcommand{\proptraj}{{\traj^{\textnormal{prop}}}}
\newcommand{\solvtraj}{{\traj^{\textnormal{solv}}}}
\newcommand{\subtask}{{g}}
\newcommand{\state}{{s}}
\newcommand{\stateref}{{s}^{\operatorname{ref}}}
\newcommand{\finalstateref}{{\stateref_{\textnormal{final}}}} %
\newcommand{\numsubtasks}{{K}}
\newcommand{\solverate}{{\hat{p}}}
\newcommand{\proposerreward}{{\reward^{\textnormal{proposer}}}}
\newcommand{\solverreward}{{\reward^{\textnormal{solver}}}}
\newcommand{\subgoalreward}{{\reward^{\textnormal{subgoal}}}}
\newcommand{\difficultyreward}{{\reward^{\textnormal{difficulty}}}}
\newcommand{\subtaskreward}{{\reward^{\textnormal{subtask}}}}
\newcommand{\diversityreward}{{\reward^{\textnormal{diversity}}}}
\newcommand{\diversityweight}{{w_{\textnormal{div}}}}
\newcommand{\actiontypes}{{\mathcal{A}}}
\newcommand{\indicator}{{\mathbb{I}}}

\newcommand{\subtaskweight}{{w_{\textnormal{sub}}}}

\newcommand{\finishtool}{{\texttt{finish}}}
\newcommand{\declaretasktool}{{\texttt{declare}}}
\newcommand{\subfinishtool}{{\texttt{subtask}}}

\definecolor{envbg}{rgb}{0.90,0.90,0.90}